\documentclass[]{article}%

\usepackage[accepted]{icml2025}
\usepackage{amsthm}

\usepackage{comment}
\usepackage{multirow}
\usepackage{natbib} % Comment and Comment out if citation is not working
\usepackage{subcaption}
\usepackage{booktabs}
\usepackage[linesnumbered,ruled,vlined,algo2e]{algorithm2e}
\RequirePackage{multirow}
\RequirePackage{footnote}
\RequirePackage{url}
\RequirePackage{amsmath}
\RequirePackage{mathrsfs}
\RequirePackage{algorithm}%
\RequirePackage{algpseudocode}%
\RequirePackage{listings}%
\RequirePackage[hidelinks]{hyperref}
\RequirePackage{crop}
\RequirePackage{graphicx}
\RequirePackage{caption}
\RequirePackage{amsmath}
\RequirePackage{array}
\RequirePackage{color}
\RequirePackage{xcolor}
\RequirePackage{amssymb}
\RequirePackage{chngpage}
\RequirePackage{totcount}
\RequirePackage{fix-cm}
\usepackage[nameinlink,capitalize]{cleveref}

\graphicspath{{Fig/}}

\theoremstyle{thmstyleone}%
\newtheorem{theorem}{Theorem}%  meant for continuous numbers
\theoremstyle{thmstyletwo}%
\theoremstyle{thmstylethree}%

\def\v{\boldsymbol}

\begin{document}

%%
% 120-word-maximum statement 
\begin{comment}
\significancestatement{BumpNet is a novel sparse, interpretable two-layer neural network architecture that leverages a sigmoidal neural network-based approach to construct basis functions, which can be adapted to solve both physics-informed machine learning and operator learning problems. The inherent sparsity of BumpNet enables efficient training with a small number of parameters and faster inference times [COMPARED TO WHAT]. BumpNet bridges neural networks with traditional numerical analysis, offering scalable, mesh-free PDE solvers for scientific and engineering applications.}
\end{comment}
%%

%\subtitle{Subject Section}
\twocolumn[
\icmltitle{BumpNet: A Sparse MLP Framework for Learning PDE Solutions}

\begin{icmlauthorlist}
\icmlauthor{Shao-Ting Chiu}{yyy}
\icmlauthor{Ioannis G. Kevrekidis}{comp}
\icmlauthor{Ulisses Braga-Neto}{yyy}
\end{icmlauthorlist}

% Affiliations (Mapping 'a' to 'yyy' and 'b' to 'comp' from the PNAS-style address block)
\icmlaffiliation{yyy}{Department of Electrical and Computer Engineering, Texas A\&M University, TX, USA}
\icmlaffiliation{comp}{Department of Chemical and Biomolecular Engineering, The Johns Hopkins University, MD, USA}

\icmlcorrespondingauthor{Shao-Ting Chiu}{stchiu@tamu.edu}
\icmlcorrespondingauthor{Ulisses Braga-Neto}{ulisses@tamu.edu}

\icmlkeywords{Machine Learning, ICML}

\vskip 0.3in
]

\printAffiliationsAndNotice{} 

\begin{abstract}
  We introduce BumpNet, a sparse multilayer perceptron (MLP) framework for PDE numerical solution and operator learning. BumpNet is based on basis function expansion, which makes them superficially similar to radial-basis function (RBF) networks. However, the basis functions in BumpNet are constructed from ordinary sigmoid activation functions in a sparse multi-layer framework. This makes BumpNet a MLP, not a RBF neural network, enabling the efficient use of modern training techniques optimized for MLPs. All parameters of the basis functions, including shape, location, and amplitude, are fully trainable. Model parsimony is encouraged through a basis function pruning scheme. BumpNet is a general meshless framework that can be combined with existing neural architectures for learning PDE solutions: here, we propose Bump-PINNs (BumpNet with physics-informed neural networks) for solving general PDEs; Bump-EDNN (BumpNet with evolutionary deep neural networks) to solve time-evolution PDEs; and Bump-DeepONet (BumpNet with deep operator networks) for PDE operator learning. We prove that BumpNets and Bump-DeepONets are universal approximators of continuous functions and continuous operators, respectively. Bump-PINNs are trained using the same collocation-based approach used by PINNs; Bump-EDNN uses a BumpNet only in the spatial domain and uses EDNNs to advance the solution in time; while Bump-DeepONets employ a BumpNet regression network as the trunk network of a DeepONet. Extensive numerical experiments demonstrate the efficiency and accuracy of BumpNets.
\end{abstract}

\section{Introduction}

Physics-Informed Neural Networks (PINNs) \cite{raissi2019a} and neural operators for PDE-related problems, such as the Deep Operator Network (DeepONet) \cite{lu_deeponet_2021}, have been demonstrated to be effective in learning the solutions of partial differential equations (PDEs), particularly due to their meshless nature and their capacity for forward and inverse modeling~\cite{karniadakis2021physics}. However, they typically rely on deep neural networks with fully-connected  layers, leading to high computational demands and limited interpretability of model behavior.

Radial-Basis Function (RBF) neural networks  provide an efficient and interpretable alternative that can overcome these limitations \cite{montazer2018radial}. RBF neural networks are typically ``shallow'', which makes them very efficient in terms of parameter count and training time. They are also interpretable, as the final trained location and shape of the learned basis functions can be observed directly. A limitation of traditional RBF neural network schemes is the use of nonlinearities based on Euclidean distance to implement the basis functions directly. This constrains the shape of the basis functions to be non-trainable and fixed. In addition, it is often the case that the centers of the basis functions are fixed, as well.

BumpNet is a novel sparse multilayer perceptron (MLP) framework for function approximation, which is based on a meshless, interpretable basis-expansion approximation, but without the drawbacks of RBF networks. BumpNets construct adaptive localized basis functions (``bumps'') from a linear combination of ordinary sigmoid nonlinearities. The ability of sigmoid MLPs to produce localized ``bumps'' was noted already in \cite{lapedes1989}. Here, we have developed a {\em weight-tying scheme} to construct each bump in such a way that their location, shape, sharpness, and orientation are fully trainable. We show that BumpNets are universal approximators of continuous functions. Model order reduction can be achieved, during training, through adaptive pruning of the bump modules.

We demonstrate that BumpNet can efficiently solve partial differential equations (PDEs) via {\em Bump-PINNs}, which use a collocation-type approach similar to PINNs \cite{raissi2019a}. Bump-PINNs have enhanced convergence speed, accuracy, and a significantly reduced parameter count in comparison with ordinary PINNs. Virtually any technique utilized to improve the accuracy of PINNs can be applied to Bump-PINNs as well. Here, we demonstrate the use of self-adaptive weights \cite{mcclenny2023self} to improve the performance of Bump-PINNs.

In time-evolution problems, Bump-PINNs can employ ``space-time'' bumps. An alternative that can be more efficient and accurate employs a Bump-PINN to approximate the spatial component in connection with a scheme to advance the solution in time, in a manner similar to finite element methods for time-evolution PDEs. Here, we propose {\em Bump-EDNN}, which trains a BumpNet neural network to represent the initial condition and uses the evolutionary neural network methodology proposed in~\cite{du2021evolutional} to advance the solution in time.

In addition, we demonstrate that BumpNet can be used in operator learning problems. In parametric PDE applications, the goal is to approximate the operator that maps between the parameter space (which may consist of equation coefficients, source terms, or initial/boundary conditions) and the solution space, in order to obtain a fast surrogate in many-query applications, such as design optimization or uncertainty quantification, when the parameter space must be sampled repetitively. DeepONets \cite{lu_deeponet_2021} are a popular and powerful neural operator architecture. Here, we propose {\em Bump-DeepONet}, which employs a BumpNet as the trunk network of a DeepONet. Bump-DeepONets accelerate convergence and reduce the overall parameter count of ordinary DeepONets, making it an efficient option for operator learning tasks. Furthermore, we show that Bump-DeepONets are universal approximators of continuous operators.

%Finally, We introduce a simple amplitude-based pruning method, which systematically removes low-amplitude bumps during training, thereby achieving parameter efficiency and enhancing training convergence. %The pruning technique can be referred to restriction process in the multigrid method.

\subsection{Related Work}

Radial basis function (RBF) methods have a long history in computational science as meshless alternatives to traditional PDE solvers \cite{belytschko_meshless_1996,bollig_solution_2012,wang2017,montazer2018radial}. PDE solving with RBF methods is typically based on a collocation approach, where the PDE residue of the basis-function approximation is minimized on a chosen set of points \cite{zhang2000,kansa1990, kansa1990a}. Radial Basis Function (RBF) neural networks were introduced in \cite{broomhead1988radial}  as a framework for multivariable functional interpolation by interpreting learning in neural networks as an interpolation task. 

 Neural networks are interpreted as adaptive basis functions expansions in \cite{cyr2020robust}, leading to a new training strategy that alternates between least squares and gradient descent (LSGD). In a similar vein, the ``Extreme Learning Machine'' (ELM) \cite{huang_extreme_2006} employs two-layer feedforward neural networks (SLFNs) that avoid iterative tuning by randomly assigning hidden node weights and analytically solving for output weights using a least-squares approach. Physics-informed radial basis neural networks have been shown to outperform conventional PINNs on problems with fine-scale features and irregular domains \cite{bai2023a}.
 
 %ELM achieves significantly faster training speeds and competitive or better generalization performance compared to traditional backpropagation and support vector machines across a variety of regression and classification tasks.%
% Robust training 
 %Along with a novel “box initialization,” this approach improves convergence and accuracy across regression and physics-informed neural network tasks, outperforming standard methods.

% SPINN
The most related methodology to the proposed BumpNet framework is  SPINN\cite{ramabathiran2021}, which is a sparse, interpretable neural network architecture for solving PDEs. Rather than purely using sigmoids to build the basis functions, SPINN still computes explicitly the distance between input and each bump center, followed by neural-network approximations to the kernel function. SPINN is not designed for operator learning. We compare Bump-PINNs and Bump-EDNN to SPINN in the experimental section. The results demonstrate that BumpNet achieves improved speed and accuracy.
%, and can be trained in space-time domain.%
% Hierearchical

Another closely related work is the hierarchical deep-learning neural network finite element method (HiDeNN-FEM) \cite{zhang2021hierarchical}. Developed independently from BumpNet, HiDeNN-FEM does construct the basis function using just neural network nonlinearities, but it is restricted to the piecewise-polynomial basis functions commonly encountered in FEM implementations, while the basis functions in BumpNet can be constructed from arbitrary sigmoids and are oriented arbitrarily in space. Another difference is that BumpNet uses basis function pruning to achieve ``h-adaptivity'', while HiDeNN-FEM achieves this by adding new basis functions. %By fixing the number of hidden layers and increasing the number of neurons (h-refinement), HiDeNN-FEM attains rh-adaptivity, further improving solution accuracy, and its flexible architecture generalizes to various interpolation functions and enrichment strategies used in advanced finite element methods.%%

\subsection{Main Contributions}

BumpNet is a novel sparse, interpretable two-layer neural network architecture that implements a basis function expansion. The primary contributions introduced by BumpNet are as follows:

\begin{enumerate}
    \item \textbf{Novel Flexible Neural Network Algorithms}: BumpNet leverages a sigmoidal neural network-based approach to construct basis functions, which can be adapted to solve both physics-informed machine learning and operator learning problems. %, demonstrating superior representational capability compared to an alternative method~\cite{ramabathiran2021}, particularly in addressing complex partial differential equation (PDE) problems %(\cref{sec:helmoltz}, \cref{sec:heat1d}). 

    \item \textbf{Pruning for h-Adaptivity}: We propose a novel simple pruning strategy that allows the basis functions to concentrate in areas of high gradient (h-adpatativity). This improves model accuracy while significantly reducing model size. %(\cref{sec:method-pruning} and \cref{sec:res-pruning}).

    \item \textbf{Sparsity for Efficient Training}: The inherent sparsity of BumpNet enables efficient training with a small number of parameters and faster inference times. %(\cref{sec:res-deeponet}).

    %\item \textbf{Solving Time-Evolution PDEs with Bump-Evolutionary Neural Networks}: BumpNet can be incorporates a neural discretization technique to enhance the evolution of neural networks. Surrogate initial function with full network, and propagate amplitude parameters with evnolutional neural network approach. This approach leverages efficiency and interpretabiliy of BumpNets, and  leads to faster, more accurate predictions on time-dependent PDEs. %(\cref{sec:res-evo}).
\end{enumerate}

\section{Methods}
\label{sec:methods}

The core of the BumpNet framework is a modular architecture that constructs basis functions (``bumps'') with regular sigmoidal neurons plus a {\em weight-tying scheme}. BumpNets are  interpretable, in the sense that there are simple formulas relating the shape, location, and amplitude of each bump in terms of the neural network weights.
%Training can be performed with any standard machine learning software. The resulting sparse neural network can perform , we can solve PDE with fewer parameters than a regular deep neural network. 
%In this section, we provide details about the BumpNet methodology.

\subsection{BumpNet Architecture}% and Weight-Tying Scheme}

\begin{figure}
    \centering
    \includegraphics[width=\linewidth]{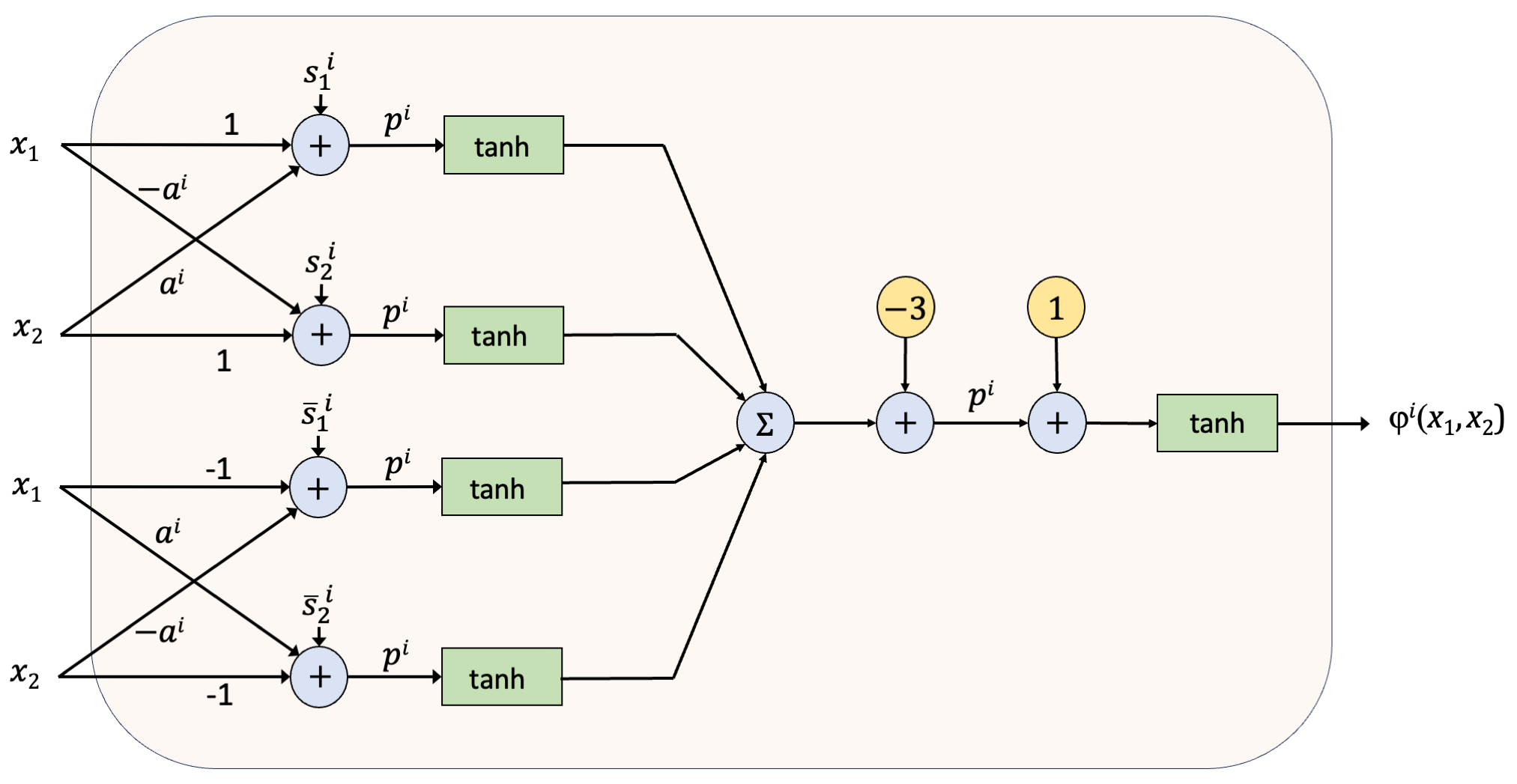}
    \caption{Network architecture of a 2D BumpNet basis module, assuming the tanh sigmoid. The output over the domain is a 2D bump, as seen in Fig.~\ref{fig-bump-vis}.}
    \label{fig-model-scheme}
\end{figure}

\begin{figure}
  \begin{subfigure}{0.50\linewidth}
    \includegraphics[width=\textwidth]{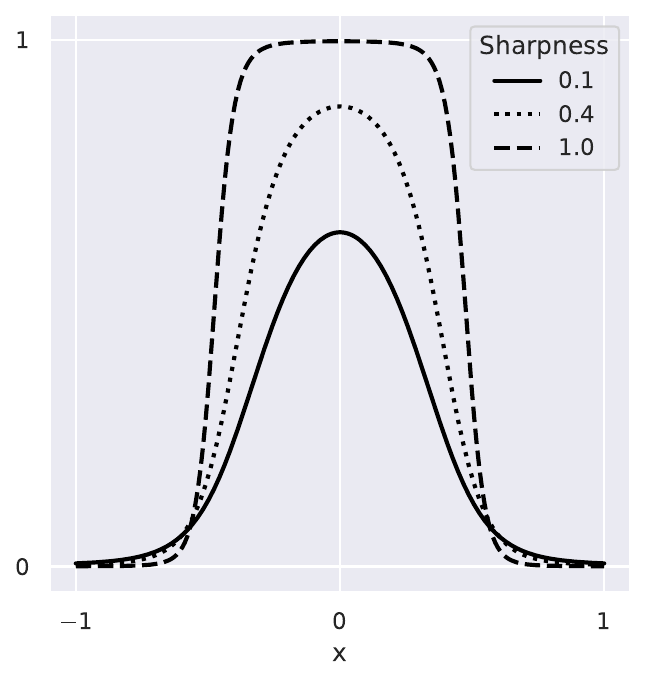}\\[-3.5ex]
  \subcaption{\label{fig:sp}}%\(BumpNet(x,y=0)\)}
  \end{subfigure}%
  \begin{subfigure}{0.50\linewidth}
    \includegraphics[width=\textwidth]{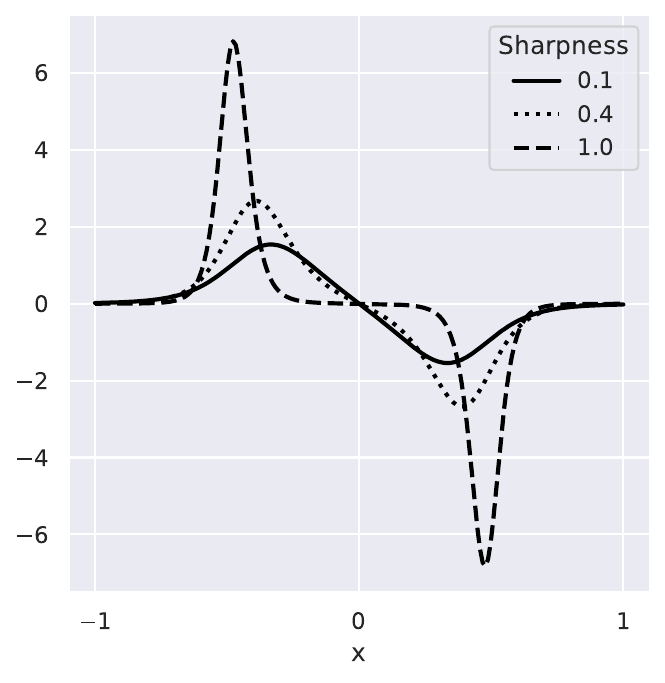}\\[-3.5ex]
  \subcaption{\label{}}%\(d BumpNet(x,y=0)/dx\)}
  \end{subfigure}%
  \\[2ex]
  \begin{subfigure}{0.50\linewidth}
    \includegraphics[width=\textwidth]{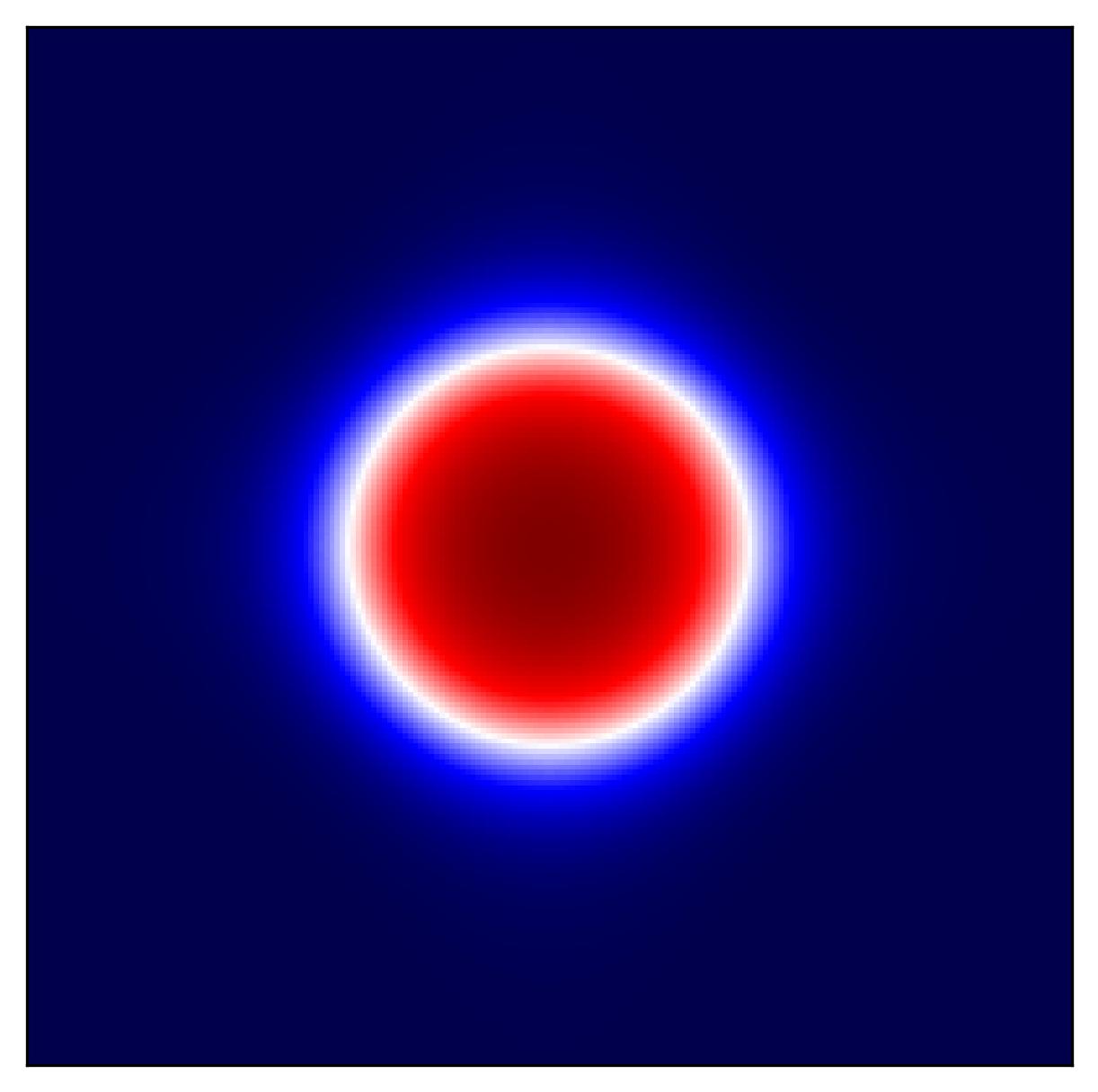}
  \subcaption{\label{fig-bump-2d}}%2D plot}
  \end{subfigure}%
  \begin{subfigure}{0.50\linewidth}
    \includegraphics[width=\textwidth]{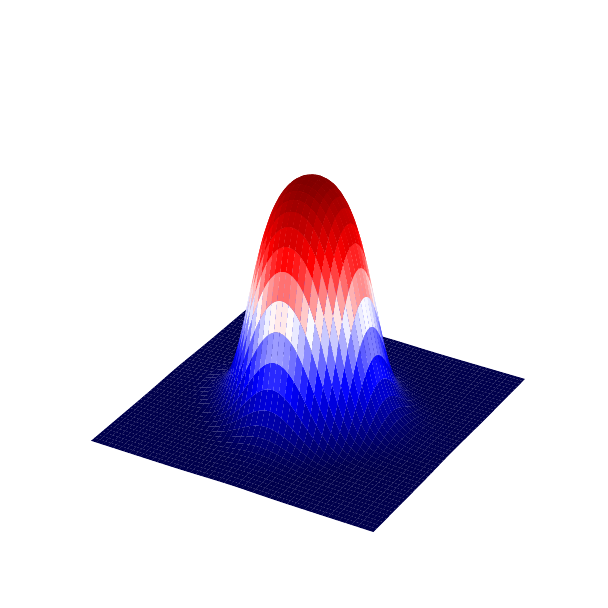}
  \subcaption{\label{fig-bump-3d}}%3D surface plot}
  \end{subfigure}%
  
  \caption{Line profile of a 2D bump in (a) and its derivative (b) for varying sharpness factor value. The bump is located the origin with unit sides. The bumps plotted in (c) and (d) have a sharpness factor of \(0.1\).} 
  \label{fig-bump-vis}
  \end{figure}

To facilitate understanding, we begin by considering the two-dimensional case, and generalize to the $n$-dimensional case in the next section. We consider the tanh sigmoid throughout (using other sigmoids requires minor changes to the architecture).
%\begin{equation}
%\tanh(x) \,=\, \frac{e^x- e^{-x}}{e^x + e^{-x}}\,, \quad -%\infty<x<\infty\,.
%\end{equation}
The output of a BumpNet for regression in the two-dimensional case is given by
\begin{equation}
  \psi(x_1,x_2) \,=\, \sum_{i=1}^m h^i\psi^i(x_1,x_2)\,,
\end{equation}
where $m$ is a hyperparameter indicating the (initial) number of basis functions in the expansion, the $h^i$ are trainable bump height parameters, for $i=1,\ldots,m$, and the $i$th basis function is given by
\begin{equation}
  \begin{aligned}
    \psi^i(x_1,x_2)&\,=\, \frac{1}{2}\tanh \left(p^i\left(\tanh\left(p^i\left(x_1+a^ix_2+s_1^{\,i}\right)\right)\right.\right.\\
        &\quad\quad+ \, \tanh\left(p^i\left(-a^ix_1+x_2+s_2^{\,i}\right)\right)\\
        &\quad\quad+ \,\tanh\left(p^i\left(-x_1-a^ix_2+\bar{s}^{\,i}_1\right)\right)\\
        &\quad\quad+\, \left.\left.\tanh\left(p^i\left(a^ix_1-x_2+\bar{s}^{\,i}_2\right)\right) -3\right)\right)+1,
  \end{aligned}
  \label{eq:get-o}
\end{equation}
% \begin{equation}
% \psi^i(x_1,x_2) \,=\, \tanh \left(p_i\left(o^{i}_{1} + o^i_{2} + \bar{o}^i_{1} + \bar{o}^i_{2} -3\right)+1\right)\,,
% \label{eq:get-o}
% \end{equation}
% where
% \begin{equation}
% \begin{aligned}
% o^{i}_{1} &\,=\, \tanh\left(p_i\left(x_1+a_ix_2+s_1^{\,i}\right)\right)\,,\\
% o^i_{2} &\,=\, \tanh\left(p_i\left(-a_ix_1+x_2+s_2^{\,i}\right)\right)\,,\\
% \bar{o}^i_{1}&\,=\, \tanh\left(p_i\left(-x_1-a_ix_2+\bar{s}^{\,i}_1\right)\right)\,,\\
% \bar{o}^i_{2}&\,=\, \tanh\left(p_i\left(a_ix_1-x_2+\bar{s}^{\,i}_2\right)\right)\,.
% \end{aligned}
% \label{eq:get-os}
% \end{equation}
corresponding to the architecture displayed in Fig.~\ref{fig-model-scheme}. The weights of this neural network are the (1) biases $s^{\,i}_{1}, s^{\,i}_{2}, \bar{s}^{\,i}_{1}, \bar{s}^{\,i}_2 \in \mathbb{R}$, (2) rotation coefficient $a^i \in \mathbb{R}$, and (3) sharpness factor $p^i  \in \mathbb{R}^+$, for $i=1,\ldots,m$. 
Note that some weights are nontrainable (e.g., the $1$ and $-1$ weights), and some weights are tied, i.e., they are mutually constrained  (e.g. the $a^i$ and $-a^i$ weights must be the negative of each other, while the ``sharpness factor'' $p^i$ is the same for all neurons). 
This architecture and weight-tying scheme guarantee that $\psi^i(x_1,x_2)$ has the shape of a two-dimensional ``bump'' over the domain. See Fig.~\ref{fig-bump-vis} for a visualization for a specific setting of the weights.

%the output $o_i$ of the $i$th module is given by the equation
%\begin{equation}
%o_i \,=\, \frac{1}{2}h_i \cdot \tanh \left(p_i\left(\sum_{j=1}^4 o_{ij} -3\right)+1\right)
%\end{equation}

% \begin{equation}
%   \begin{bmatrix} o_1 \\ o_2 \\ o_3 \\ o_4 \end{bmatrix} \,=\, \sigma_1\left(\begin{bmatrix}1 & a_i\\ -a_i & 1 \\ -1 & -a_i \\ a_i & -1\end{bmatrix}%
%   \begin{bmatrix}x_1 \\ x_2\end{bmatrix} + 
%  \begin{bmatrix}s^{i}_{1} \\ s^{i}_{2} \\ s^{i}_{3} \\ s^{i}_{4}\end{bmatrix}\right)
%  \label{eq-input-layer}
% \end{equation}
% and the activation functions are 
% \begin{align}
%   \sigma_1(x) &= \tanh(s_p \cdot x),\\
%   \sigma_2(x) &= \frac{1}{2} \tanh(s_p (x - 3) + 1).
% \end{align}

% Each activation function incorporates a trainable scaling factor, $s_p$, which adjusts the sharpness of transitions at boundaries (\cref{fig-model-scheme}). In the second layer, we modify the activation function by applying an offset of 3, ensuring that activation is restricted to a specific region inside the designated rectangle. This mechanism can be understood as enforcing geometric constraints through four inequalities, which define a rectangular activation region (\cref{eq-ineq}).
The four neurons in the first layer in Fig.~\ref{fig-model-scheme} define a rectangular support for the bump that is the intersection of the four half-planes defined by
\begin{equation}
\begin{aligned}
  H^i_1 &: x_1 + a^i x_2 + s^i_1 \geq 0\,,\\
  H^i_2 &: -a^i x_1 + x_2 + s^i_2 \geq 0\,,\\
  \bar{H}^{\,i}_1 &: -x_1 -a^i x_2 + \bar{s}^i_1 \geq 0\,,\\
  \bar{H}^{\,i}_2 &: a^i x_1 - x_2 + \bar{s}^i_2 \geq 0\,.
  %O &= h \cdot \text{Bool}(O_1 \cap O_2 \cap O'_1 \cap O'_2),
\end{aligned}
\label{eq-ineq4}
\end{equation}
Note that the boundaries of $H_1$ and $\bar{H}_1$ are parallel to each other, and similarly for $H_2$ and $\bar{H}_2$. It is easy to see that the the rectangular support sides are given by
\begin{equation}
\begin{aligned}
l_1^i &\,=\, \frac{s^{\,i}_1 + \bar{s}^{\,i}_1}{\sqrt{(a^i)^2 +1}}\,,\\
l_2^i &\,=\, \frac{s^{\,i}_2 + \bar{s}^{\,i}_2}{\sqrt{(a^i)^2 + 1}}\,,
\end{aligned}
\label{eq-bump-geo1}
\end{equation}
while the bump center coordinates are
\begin{equation}
\begin{aligned}
c_1^i &\,=\, \frac{(\bar{s}^{\,i}_1 - s^{\,i}_1) -a^i(\bar{s}^{\,i}_2 - s^{\,i}_2)}{2((a^i)^2 + 1)}\,,\\
c_2^i &\,=\, \frac{a_i (\bar{s}^{\,i}_1 - s^i_1) + (\bar{s}^{\,i}_2 - s^{\,i}_2)}{2((a^i)^2 + 1)}\,.
\end{aligned}
\label{eq-bump-geo2}
\end{equation}
The interpretability of BumpNet comes from the fact that the support sizes, centers, orientation, sharpness, and height of each bump can be read from the trained BumpNet directly, as seen above. 

\def\n{n}  
\subsection{Generalization to $\n$ Dimensions}

In $\n$ dimensions, the output of a BumpNet for regression is
\begin{equation}
  \psi(\v{x})\,=\, \sum_{i=1}^m h^i\psi^i(\v{x})\,,
\label{eq-Bump1}
\end{equation}
for $\v{x} \in R^\n$, where $m$ is the number of basis functions and
% \begin{equation}
%   \begin{aligned}  
%     \psi^i(\v{x}) &\,=\, \frac{1}{2}\left(1+\tanh\! \left(\!p^i \!\left( \sum_{j=1}^\n\tanh\!\left(p^i(\v{x}^T\! \v{\beta}^i_j+ s^i_j)\right) \right.\right.\right.\\
%         &\quad\:\:+\! \sum_{j=1}^\n \tanh\!\left(p^i(-\v{x}^T\! \v{\beta}^i_j + \bar{s}^{\,i}_j)\right)-(2n-1)\!\Bigg)\!\Bigg)\!\Bigg).
%   \end{aligned}
%   \label{eq:get-on}
% \end{equation}
\begin{equation}
  \begin{aligned}  
    \psi^i(\v{x}) &\,=\, \mathrm{squash}\left(\sum_{j=1}^\n\tanh\!\left(p^i(\v{x}^T\! \v{\beta}^i_j+ s^i_j)\right)\right.\\
        &\quad\:\:+\! \sum_{j=1}^\n \tanh\!\left(p^i(-\v{x}^T\! \v{\beta}^i_j + \bar{s}^{\,i}_j)\right)-(2n-1)\!\Bigg).
  \end{aligned}
\label{eq-Bump2}
\end{equation}
where
\begin{equation}
\mathrm{squash}(x) \,=\, \frac{1}{2}(1+\tanh(p_i x))\,.
\label{eq-Bump3}
\end{equation}
% \begin{equation}
%   \begin{aligned}  
%     \psi^i(\v{x}) &\,=\, \frac{1}{2}\tanh\! \left(\!p^i \!\left( \sum_{j=1}^\n\tanh\!\left(p^i(\v{x}^T\! \v{\beta}^i_j+ s^i_j)\right) \right.\right.\\
%         &\quad\:\:+\! \sum_{j=1}^\n \tanh\!\left(p^i(-\v{x}^T\! \v{\beta}^i_j + \bar{s}^{\,i}_j)\right)-(2n-1)\!\Bigg)\!\Bigg)\!+1.
%   \end{aligned}
%   \label{eq:get-on}
% \end{equation}
Hence, there are now $2\n$ neurons in the first layer of each BumpNet module, with weights $(\v{\beta}^i_j,s^i_j)$ and $(-\v{\beta}^i_j,\bar{s}^{\,i}_j)$, $j=1,\ldots,\n$. In order to determine the weight-tying scheme for the direction vectors $\v{\beta}^i_j$, 
%There are simple formulas relating the BumpNet weights and the support size and bump center coordinates. First,
note that we need $n$ orthogonal pairs of half-spaces to define the support of the $i$th bump. The $j$th pair of half-spaces is: 
% \begin{align}
%   H_1 &: x_1 + a_1 x_2 + a_2 x_3 + \cdots + a_{n-1}x_n + s_1 \geq 0\,,\\
%   \bar{H}_1 &: -x_1 - a_1 x_2 - a_2 x_3 + \cdots - a_{n-1}x_n + \bar{s}_1 \geq 0\,,
% \label{eq-ineq4}
% \end{align}
\begin{equation}
\begin{aligned}
  H^i_j &: \v{x}^T\! \v{\beta}^i_j + s^i_j\geq 0\,,\\
  \bar{H}^{\,i}_j &: -\v{x}^T\! \v{\beta}^i_j+ \bar{s}^{\,i}_j \geq 0\,.
\end{aligned}
\label{eq-ineq5}
\end{equation}
We can make the first pair of half-spaces be oriented in an arbitrary direction by letting $\v{\beta}^i_1 = [1, a^i_1, \dots, a^i_{\n-1}]^T$, where $a^i_1,\ldots,a^i_{\n-1}$ are arbitrary real numbers. The remaining half-space pairs can be determined by using the constraint that they must be orthogonal to the first pair and orthogonal to each other. One way to do this is to find values of $\v{\beta}_j$ using the Gram-Schmidt orthogonalization process on the full-rank $\n \times \n$ matrix $[\v{\beta}^i_1, \v{e}_2, \ldots \v{e}_\n]$, where $\v{e}_j$ has $1$ in the $j$th position and $0$ everywhere else. The resulting orthogonal matrix $B^i = [\v{\beta}^i_1,\ldots,\v{\beta}^i_\n]$ contains the desired coefficients in its columns. 
%The only free parameters are $a^i_1,\ldots,a^i_{n-1}$, $s^i_1,\ldots,s^i_n$, and $\bar{s}^{\,i}_1,\ldots,\bar{s}^{\,i}_n$, the number of which grows linearly with $n$. 

As an example, with $\n=2$, we have $\v{\beta}^i_1 = [1, a^i]^T$ and the Gram-Schmidt orthogonalization process produces 
\begin{equation}
\begin{bmatrix}
    \begin{bmatrix}
        1 \\ a^i
    \end{bmatrix},
    \begin{bmatrix}
        0\\1
    \end{bmatrix}
\end{bmatrix} \:\:\rightarrow\:\: 
\begin{bmatrix}
        \begin{bmatrix}
            1\\a^i
        \end{bmatrix},
        \begin{bmatrix}
            -a^i \\ 1
        \end{bmatrix}
    \end{bmatrix}
\end{equation}
so that $\v{\beta}^i_2 = [-a^i,1]^T$. This gives the coefficients in \eqref{eq:get-o} and Fig.~\ref{fig-model-scheme}. With $\n=3$, we would have $\v{\beta}^i_1 = [1, a^i_1, a^i_2]^T$ and the Gram-Schmidt orthogonalization process produces 
\begin{equation}
\begin{bmatrix}
    \begin{bmatrix}
        1 \\ a^i_1\\ a^i_2
    \end{bmatrix},
    \begin{bmatrix}
        0\\1\\0
    \end{bmatrix},
    \begin{bmatrix}
        0\\0\\1
    \end{bmatrix}
\end{bmatrix} \:\:\rightarrow\:\: 
\begin{bmatrix}
        \begin{bmatrix}
            1\\a^i_1\\a^i_2
        \end{bmatrix},
        \begin{bmatrix}
            -a^i_1\\ (a^i_2)^2 + 1\\- a^i_1 a^i_2
        \end{bmatrix}
        \begin{bmatrix}
            -a^i_2\\ 0 \\ 1
        \end{bmatrix}
    \end{bmatrix}
\end{equation}
so that $\v{\beta}^i_2 = [-a^i_1,(a^i_2)^2+1,-a^i_1a^i_2]^T$ and $\v{\beta}^i_3 = [-a^i_2,0,1]^T$. It should be clear how to modify the first-layer of the BumpNet module in Fig.~\ref{fig-model-scheme} in the $n=3$ case, and beyond.

It can be shown that the generalization of \eqref{eq-bump-geo1} to find the support size in the $\n$-dimensional case is
\begin{equation}
  l^i_j \,=\, \frac{s^i_j + \bar{s}^{\,i}_j}{||\v{\beta}^i_j||} \,, \quad j=1,\ldots,\n,
\label{eq-bump-geon1}
\end{equation}
where $||\cdot||$ denotes the Euclidean norm, while the generalization of \eqref{eq-bump-geo2} to obtain the bump center coordinates is
\begin{equation}
    \begin{bmatrix}
        c^i_1 \\ \cdots\\ c^i_\n
      \end{bmatrix}\,=\, \left(B^i\right)^{-1}
\begin{bmatrix}
  \bar{s}^{\,i}_1 - s^i_1  \\ \cdots\\ \bar{s}^{\,i}_\n - s^i_\n
      \end{bmatrix}      \,.
\label{eq-bump-geon2}
\end{equation}

\subsection{Universal Function Approximation Property}

BumpNets are universal approximators of continuous functions on compact domains, as shown next. The proof can be found in the Appendix.

\begin{theorem}[Universal Approximation of Continuous Functions by BumpNet]
Let $\Omega \subset R ^d$ be a compact physical domain. The class of BumpNets defined by equations \eqref{eq-Bump1}--\eqref{eq-Bump3} is dense in $C(\Omega)$ with respect to the uniform norm. That is, for any $f \in C(\Omega)$ and any $\varepsilon > 0$, there exists a BumpNet $\psi$ such~that
\[
\| f(\v{x}) - \psi(\v{x})\| < \varepsilon\,, \:\: \textrm{for all} \:\: \v{x} \in \Omega.
\]
\label{thm:uni-bump}
\end{theorem}

\subsection{Detailed Training Procedure}
\label{sec:detailed}

As can be seen in \eqref{eq-bump-geon1}, in order to obtain a nonvanishing bump support, we must have $s^i_j + \bar{s}^{\,i}_j >0$, for $j=1,\ldots,\n$. To achieve this, rather than training the biases $s^i_j$ and $\bar{s}^{\,i}_j$ directly, we employ a reparametrization. First, we introduce trainable parameters {\color{black} $W^i_j\in R$} such that
\begin{equation}
l^i_j \,=\, \exp\left(W^i_j\right) >\,0 \,,\:\:\: j=1,\ldots,\n\,.
\label{eq-train-size}
\end{equation}
In addition, in order to keep the center coordinates $(c^i_1,\ldots,c^i_\n)$ in \eqref{eq-bump-geon2} of each bump in or around the PDE domain, we introduce trainable parameters $v^i_{j} \in R$ such that
\begin{equation}
c^i_j \,=\, x_{jl} + \frac{1}{2}(\tanh(v^i_{j}) + 1) (x_{jr} - x_{jl})\,, \:\:\: j=1,\ldots,\n\,,
\label{eq-train-center}
\end{equation}
where $(x_{1l}, x_{1r})\times\cdots\times (x_{\n l}, x_{\n r})$ is the bounding box around $\Omega$, with $x_{jl} < x_{jr}$, for $j=1,\ldots,\n$. If the domain $\Omega$ is hyperrectangular, which is often the case, then \eqref{eq-train-center} constrains the bumps to be inside the domain. This enables BumpNet to effectively model complicated functions, such as the solution of the Helmholtz PDE (as is shown in Section \ref{sec:helmoltz}).

Once the parameters $W^i_j$ and $v^i_j$ are updated at each iteration of gradient descent, the values of $l^i_j$ and $c^i_j$ are computed via \eqref{eq-train-size} and \eqref{eq-train-center}, respectively, and the updated values of the biases $s^i_j$ and $\bar{s}^{\,i}_j$ can be obtained by solving the system of equations given by \eqref{eq-bump-geon1} and \eqref{eq-bump-geon2}. This system can be solved quite easily to give:
\begin{equation}
\begin{aligned}
        s^i_j& \,=\, \frac{1}{2}\, l^i_j\,||\v{\beta}^i_j|| - \frac{1}{2}\sum_{k=1}^\n c^i_k \v{\beta}^i_k\,, \:\:\: j=1,\ldots,\n\,,\\
        \bar{s}^{\,i}_j& \,=\, \frac{1}{2}\,l^i_j\,||\v{\beta}^i_j|| + \frac{1}{2}\sum_{k=1}^\n c^i_k \v{\beta}^i_k \,, \:\:\: j=1,\ldots,\n\,.
\end{aligned}        
\label{eq-bump-geon3}
\end{equation}
The remaining weights in the neural network are trained as usual. The procedure for training a BumpNet for regression is summarized in Algorithm~\ref{Alg1}.

%\begin{comment}
\begin{algorithm}[H]
%\DontPrintSemicolon
\SetKwFunction{FNet}{Bump}
\SetKwFunction{FLoss}{Loss}
\SetKwFunction{FGrad}{Grad}
\SetKwFunction{FUpdate}{Update}
\SetKwInput{KwParam}{Trainable Parameters}
\SetKwInput{HyperKwParam}{Hyperparameters}
\SetKwInput{TrainParam}{Trainable Parameters}
\SetKwInput{Input}{Input}
\SetKwInput{Initialize}{Initialization}
\SetKwInput{Train}{Training}
\SetKwInput{Output}{Output}
\SetKwInput{Hyper}{Hyperparameters}

\Input{Training data $\{(\v{x}_{1},y_1),\ldots, (\v{x}_{N},y_N)\}$;}
\HyperKwParam{Number of bumps $m$, learning rate $\eta$, number of iterations $T$;}
\KwParam{$\v{\theta} = \{a^i_j,W^i_{j},v_{j}^{i},p^i,h^i\}$;}
\Initialize{Using Eqs.~\ref{eq-train-size} and \ref{eq-train-center}, initialize $\{a^i_j,W^i_j,v^i_j\}$ such that the bumps are uniformly distributed over the domain bounding box and parallel to its boundaries. Initialize $h_i = 1$ and $p_i = \log 3.5$ (all bumps have unit height with a sharp profile);}
\Train{\\
  \For{$t = 1$ \KwTo $T$}{
    \For{$i = 1$ \KwTo $m$}{
          \For{$j = 1$ \KwTo $n$}{
            Compute size $l^i_j$ using \eqref{eq-train-size};\\
            Compute center $c^i_j$ using \eqref{eq-train-center};\\
            Compute biases $s^i_j$ and $\bar{s}^{\,i}_j$ using \eqref{eq-bump-geon3};
              }}
  Compute loss $L(\v{\theta}) \gets \frac{1}{N} \sum_{i=1}^N (\psi(\v{x}_k)-y_k)^2$;\\
  
  Compute gradient $g \gets \FGrad(L(\v{\theta}))$;\\

  Update trainable weights $\v{\theta} \gets \FUpdate(\v{\theta}, g, \eta)$;
}
}
\caption{Training BumpNet for Regression.}
\label{Alg1}  
\end{algorithm}
%\end{comment}

\subsection{BumpNet Pruning}\label{sec:method-pruning}

Model reduction can be achieved, with minimal loss of accuracy, by removing non-essential bumps. A simple way to do this, which we adopt here, is to prune bumps for which the height parameter $h_i$ falls below a certain pre-specified threshold hyperparameter $q\!>\!0$. This is done dynamically, after each $r$ training steps, where $r$ is a hyperparameter. This dynamical pruning process can improve the convergence speed and accuracy of the resulting BumpNet.
%After each $m$ training steps, a fraction $f$ of the shortest bumps are removed, where $m$ and $f$ are user-specified hyperparameters.

\subsection{Bump-PINNs}

Consider a general PDE:
\begin{equation}
  \begin{aligned}  
    \mathcal{N}_{\v{x}}[u(\v{x})]  &\,=\, f(\v{x}), \:\:\:\v{x} \in \Omega\,,\\
    \mathcal{B}_{\v{x}}[u(\v{x})] &\,=\, g(\v{x}), \:\:\:\v{x} \in \partial \Omega\,,
  \end{aligned}
\end{equation}
where $\v{x}$ may include space, time, and other physical variables, $\mathcal{N}$ is a differential operator, $\mathcal{B}$ is a boundary operator, $u$ is the solution of rgw PDE, $f$ is the source function, $g$ is the boundary condition, and $\Omega$ is the domain of the PDE. 

BumpNets can be trained to satisfy a PDE and approximate its solution, in the manner of physics-informed neural networks (PINN) \cite{raissi2019a}. The resulting algorithm is called Bump-PINN. Let $\hat{u}(\v{x}; \v{w})$ denote the neural network, where $w$ denotes the network weights. By applying Automatic Differentiation to calculate differential terms, the network is trained to minimize the loss function: 
\begin{equation}
  \begin{split}
      \mathcal{L}(\v{w}) 
      &= \frac{1}{N_r}\sum_{i=1}^{N_r}\left|N_{\v{x}}[\hat{u}(\v{x}^r_i; \v{w})] - f(\v{x}^r_i)\right|^2\\
      &+ \frac{1}{N_b} \sum_{i=1}^{N_b} \left|\mathcal{B}_{\v{x}}[\hat{u}(\v{x}^b_i; \v{w})] - g(\v{x}^b_i)\right|^2
  \end{split}
\end{equation}
where $\{\v{x}^r_i\}_{i=1}^{N_r} \subset \Omega$ and $\{\v{x}^b_i\}_{i=1}^{N_b} \subset \partial \Omega$ are residual and boundary points, respectively. 

\subsection{Bump-SAPINNs}

Weighting schemes can be applied to the PINN loss function in order to improve convergence speed and accuracy. Here, we consider the Self-Adaptive PINN (SAPINN) weighting scheme of~\cite{mcclenny2023self}, resulting in the Bump-SAPINN method (self-adaptive Bump-PINN). The original SAPINN requires full-batch training to update the self-adaptive weights, which can impose substantial memory and computational demands (although a modification to allow stochastic mini-batch training is also proposed in~\cite{mcclenny2023self}). In contrast, the proposed BumpNet architecture is highly efficient, with a parameter count that is 20 to 100 times smaller than that of standard multi-layer perceptrons (MLPs). Despite its compact design, BumpNet demonstrates comparable performance to SAPINN in leveraging self-adaptive weights, achieving similar accuracy in solution representation, as will be seen in Section~3. Moreover, BumpNet significantly reduces training time due to its lightweight architecture and efficient parameterization. These results highlight the advantages of BumpNet in terms of memory efficiency, computational speed, and scalability, making it a promising alternative to traditional PINN architectures for self-adaptive frameworks.

\subsection{Bump-EDNN}

To solve time-evolution PDEs, Bump-PINNs can be employed with bumps over the entire space-time domain. Here, we consider an alternative, namely, the recently-proposed evolutional deep neural networks (EDNN) \cite{du2021evolutional}. Consider a time-evolution PDE of the form
\begin{equation}
\begin{aligned}
  \frac{\partial u(\v{x},t)}{\partial t} &\,=\, \mathcal{N}_{\v{x}}[u(\v{x},t)]\, , \:\:\: \v{x} \in \Omega\,, \:t \in (0,T)\,,\\
    u(\v{x},0) & \,=\, h(\v{x})\,, \:\:\: \v{x} \in \Omega\,,\\
    u(\v{x},t)&\,=\, g(\v{x},t)\,, \:\:\:\v{x} \in \partial \Omega\,, \:t \in (0,T)\,.
  \end{aligned}
\end{equation}
A Bump-EDNN $\hat{u}(\v{x},\v{w}(t))$ first determines an initial value for the weights $\v{w}(t=0)$ by fitting the initial condition $h(\v{x})$ with a regular BumpNet, and then advances the solution by updating the weights $\v{w}(t)$ using an estimate of the derivative $d \v{w}(t)/dt$, while enforcing the lateral boundary conditions automatically using the method in \cite{lu2021physics}. Therefore, the Bump-EDNN approximates the solution with spatial bumps at each different times $t$, where the bumps change with $\v{w}(t)$, but without further gradient descent training. To obtain the estimate of $d \v{w}(t)/dt$, first note that replacing $\hat{u}(\v{x};\v{w}(t))$ in the PDE, one obtains, by the chain rule of differentiation,
\begin{equation}
  \frac{\partial \hat{u}(\v{x};\v{w}(t))}{\partial t}\,=\, \frac{\partial \hat{u}(\v{x};\v{w}(t))}{\partial \v{w}(t)} \cdot \frac{d \v{w}(t)}{d t} \,=\, \mathcal{N}_{\v{x}}[\hat{u}(\v{x};\v{w}(t))]\,.
\end{equation}
Hence, an approximation of the derivative at each time $t$ can be obtained as
\begin{equation}
    \frac{d\v{w}(t)}{d t} \,\approx\, \arg\min_{\v{\gamma} \in {\mathcal W}} \: \sum_{i=1}^{N_w}\! \left(\!\frac{\partial \hat{u}(\v{x}^w_i;\v{w}(t))}{\partial \v{w}(t)}  \cdot \v{\gamma} - \mathcal{N}_{\v{x}}[\hat{u}(\v{x}^w_i;\v{w}(t))]\!\right)^{\!2}\!\!, %}_{J(\gamma)}
    \label{eq-ednn-govern}
  \end{equation}
  where ${\mathcal W}$ is the weight space, $\{\v{x}_i^w\}_{i=1}^{N_w}$ is a specified set of collocation points covering the domain $\Omega$, and all derivatives on the right-hand side can be computed by automatic differentiation. This minimization problem \eqref{eq-ednn-govern} can be solved using any standard optimization technique; here, we simply use the method proposed in \cite{du2021evolutional}, which is to solve the first-order optimal necessary conditions. Given the initial weight value $\v{w}(t=0)$ and $d \v{w}(t)/dt$, the evolution of $\v{w}(t)$ can be used any integration method, such as simple forward Euler method. Here, we use 3rd-order Runge-Kutta scheme for obtaining both good accuracy and computational speed.
  %The optimization of \cref{eq-ednn-govern} forms a normal equation $J^T J \gamma = J^T N$ where $J$ is $n_u \times n_\theta$ of collocation points $N_u$ and network parameters $N_\theta$. This can be solved by conjugate gradient method. We set tolerance to $1e-8$.

The advantage of this approach for solving time-dependent PDEs is that the network only needs to be trained on the initial condition. Time propagation is handled by integration, eliminating the need for retraining at each time step. However, one still needs to solve the optimization problem \eqref{eq-ednn-govern} at each new time step. 
To decrease the computational effort, only a subset of the weights $\v{w}(t)$ of the BumpNet can be included in the computation of \eqref{eq-ednn-govern} and updated during time-stepping; here, only the height parameters $h_i$ are updated. Thus, the bump shapes and locations are determined during training with the initial condition and kept fixed and subsequently only the coefficients that multiply each bump are updated. Another way to reduce the computational effort is to prune the smallest bumps after the BumpNet is trained on the initial condition, leading to a sparser basis expansion and a smaller neural network. All of this mirrors the procedure used in classical basis-expansion methods for solving PDEs, such as the Proper Orthogonal Decomposition (POD) \cite{benner2015survey}.
  
\subsection{Bump-DeepONet}

%DeepONets approximate operators between infinite-dimensional spaces. 

\begin{figure}
    \centering
    \includegraphics[width=\linewidth]{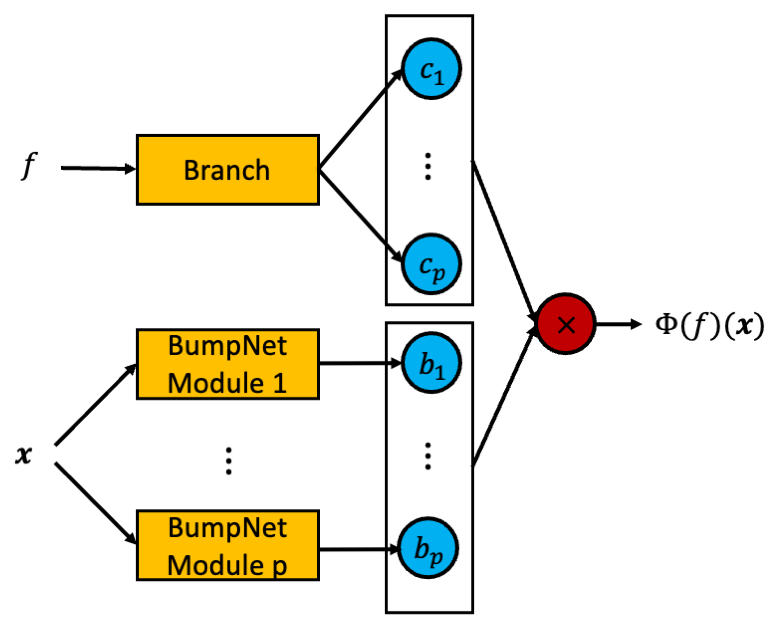}
    \caption{Network architecture of a Bump-DeepONet.}
    \label{fig-Bump_DeepONet}
\end{figure}

DeepONets \cite{lu_deeponet_2021} are neural approximators of operators between function spaces. In the case of a PDE solution operator, the input function is a parameter and the output function is the corresponding PDE solution. In a DeepONet, there are ``branch'' and ``trunk'' neural networks. The branch network takes the input parameter function $f$ discretized a $m$ locations, $[f(\v{x}_1), \ldots,f(\v{x}_m)]$, and produces $p$ scalar outputs $b_1,\ldots,b_p$. On the other hand, the trunk network takes the input coordinate $\v{x}$ where the solution is to be evaluated and also outputs $p$ scalar values $t_1,\ldots,t_p$. The output of the DeepONet is obtained by taking the dot product of these outputs:
\begin{equation}
    \Phi(f(\v{x}_1), \ldots,f(\v{x}_m))(\v{x}) \,=\, \sum_{i=1}^p b_{i}(f(\v{x}_1), \ldots,f(\v{x}_m)) \cdot t_i(\v{x})\,.
    \label{eq:deeponet}
\end{equation}
(Sometimes a bias term is added, which can improve accuracy.) In Bump-DeepONet, the branch network is a regular MLP, while the trunk network consists of $p$ BumpNets that compute the basis outputs $t_1(\v{x}),\ldots,t_p(\v{x})$. In practice, each of these Bumpnets may consist of a single BumpNet module. See Figure~\ref{fig-Bump_DeepONet} for an illustration.

% This can be thought as a basis expansion approximation, where the trunk network computes adaptive basis functions, and the branch network computes the expansion coefficients. The bias term $b_0$, while not strictly necessary, often enhances performance. 

% \subsection{Imposing Dirichlet boundary condition}

% Any regular PINN can strictly fulfill Dirichlet boundary condition on the input~\cite[Sec. 12.3]{braga-neto_fundamentals_2024}

% \begin{equation}
% \begin{split}
%     u_{Dirichlet}(x) &= (1 - e^{x_{1_b} - x_1})(1 - e^{x_1 -x_{1_a}} )\\
%                      &(1 - e^{x_{2_b} - x_2}) (1 - e^{x_2 - x_{2_a}}) u(\v{x})
% \end{split}
% \label{eq:drichlet}
% \end{equation} 

% Applying Dirichlet boundary conditions ensures that the solution adheres to fixed values at the domain boundaries, which is essential for problems requiring strict constraints, such as heat equation (\cref{sec:heat1d}). By incorporating these conditions directly into the neural network architecture, the model avoids the need for additional penalty terms in the loss function, leading to more stable and accurate training.

\subsection{Universal Operator Approximation Property}

%\section{Universal Approximation Property}\label{sec:results}

% BumpNets are universal approximators of continuous functions on compact domains, as shown next. The proofs are in the Appendix.

% \begin{theorem}[Universal Approximation of Continuous Functions by BumpNet]
% Let $\Omega \subset R ^d$ be a compact physical domain. The class of BumpNets defined by equations \eqref{eq-Bump1}--\eqref{eq-Bump3} is dense in $C(\Omega)$ with respect to the uniform norm. That is, for any $f \in C(\Omega)$ and any $\varepsilon > 0$, there exists a BumpNet $\psi$ such~that
% \[
% \| f(\v{x}) - \psi(\v{x})\| < \varepsilon\,, \:\: \textrm{for all} \:\: \v{x} \in \Omega.
% \]
% \end{theorem}

Bump-DeepONet are universal approximators of continuous operators
on compact function spaces, as shown next. The proof is in the
Appendix.

\begin{theorem}[Universal Approximation of Continuous Operators by Bump--DeepONet]
Let $K$ and $\Omega$ be compact sets in $R^p$ and $R^d$,
   respectively, and let $\mathcal{P}$ be a compact set in $C(K)$ under the uniform norm. If $G:\mathcal{P} \rightarrow C(\Omega)$ is any continuous operator, then for
   any $\varepsilon > 0$, there exists a Bump-DeepONet $\Phi$ and points $\v{x}_1,\ldots,\v{x}_m \in K$, such that
\[
\sup_{\substack{f \in \mathcal{P}\\\v{x}\in\Omega}}
\left|G(f)(\v{x}) - \Phi(f(\v{x}_1), \ldots,f(\v{x}_m))(\v{x}) \right| \,<\, \varepsilon .
\]
\end{theorem}

\section{Experimental Results}\label{sec:results}

In this section, we employ well-known PDE benchmarks to investigate the performance of Bump-PINN, Bump-EDNN, and Bump-DeepONet, comparing them to alternative methods in the literature.

\subsection{Bump-PINN Performance}\label{sec:pde2d}

We evaluate the baseline Bump-PINN architecture against the baseline PINN \cite{raissi2019a}, self-adaptive PINN \cite{mcclenny2023self}, Bump-SAPINN (that is, a Bump-PINN with self-adaptive weights), as well as SPINN \cite{ramabathiran2021}, a method similar to BumpNets. Our evaluation considers accuracy (relative L1 error against the analytical solution), training time, and parameter count across a number of well-known PDE benchmarks. We use a exponential learning rate scheduler with scale 0.9 for every 1000 training steps. 

%We employ $10^3$ collocation points and $200$ boundary points in each case. 
%Gradient descent is implemented with the Adam optimizer, with a customized exponential schedule for the learning rate.

%\subsubsection{Benchmarks}

\subsubsection{Inhomogeneous Helmholtz Equation}
\label{sec:helmoltz}
\begin{equation}
    u_{xx}(x,y) + u_{yy}(x,y) + k^2 u(x,y) \,=\, q(x,y)\,, 
\label{eq-helmotz-prob}
\end{equation}
where $(x,y) \in [-3,3]^2$, $k=1$, with zero boundary conditions. The forcing term $q(x,y)$ is manufactured to give the solution
\begin{equation}
    u(x,y) = \sin(\pi x) \sin(\pi y)\,.
\end{equation}
The Bump-PINN is initialized with $10 \times 10$ bumps uniformly distributed over the domain $(x,y)$ plane, and connected seamlessly. For training, a random batch of $10^4$ collocation points within the domain $[-3,3]^2$ and $200$ uniformly distributed boundary points on each boundary are sampled. The network is trained using the Adam optimizer, with an initial learning rate of 7e-2 for Bump-PINN and Bump-SAPINN and 1e-3 for PINN and SAPINN. 

\begin{figure}[h]
  \begin{subfigure}{\linewidth}
      \includegraphics[width=\linewidth]{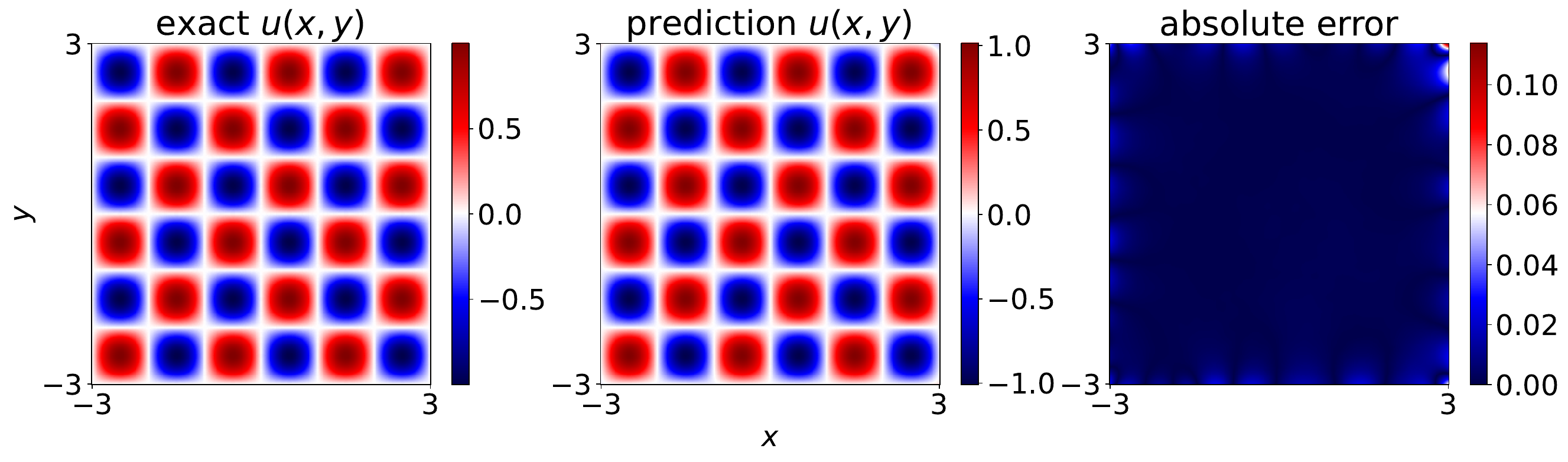}
      \subcaption{(\textit{Left}) Exact solution. (\textit{Middle}) Prediction of the BumpNet-PINN. (\textit{Right}) Absolute difference.}
  \end{subfigure}
  \begin{subfigure}{\linewidth}
      \includegraphics[width=\linewidth]{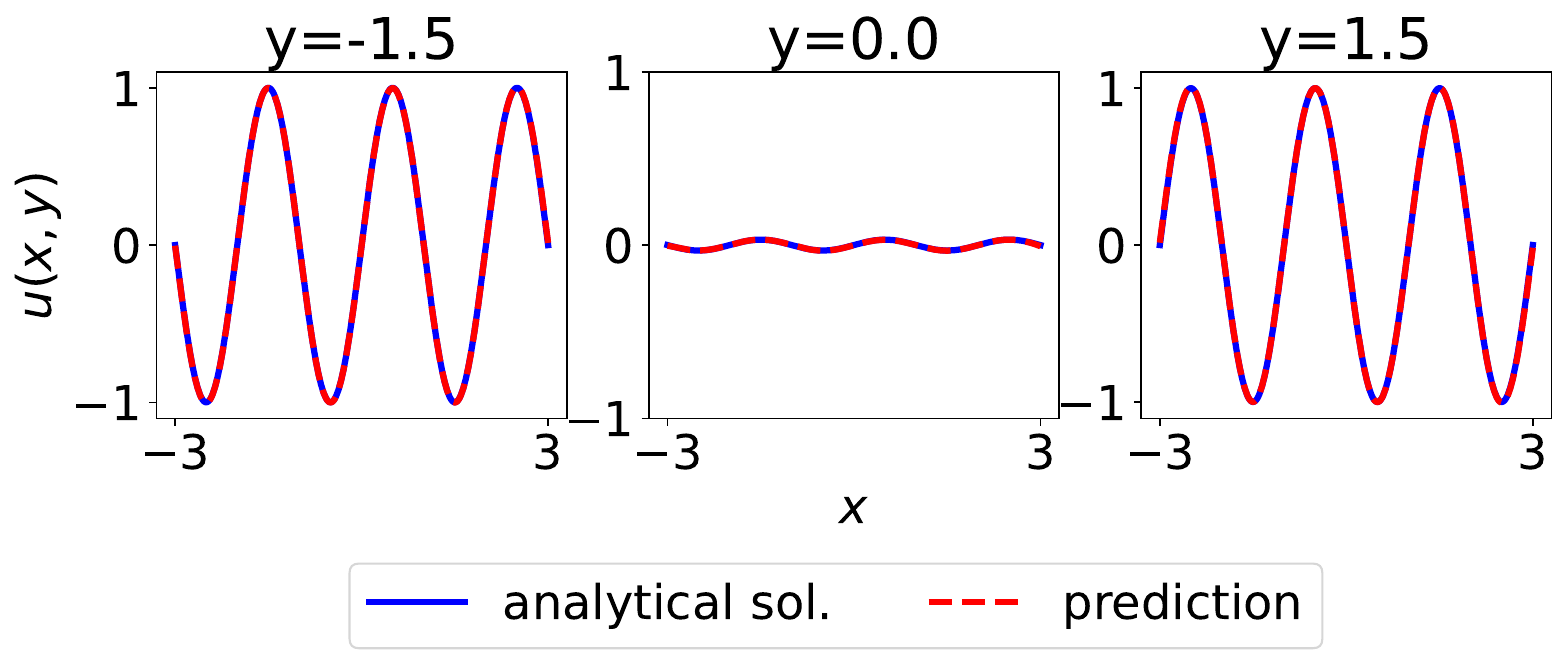}
      \subcaption{Comparison of prediction and exact solution on $y= -1.5, 0.0, 1.5$ over $x\in [-3,3]$.}
  \end{subfigure}
  \caption{Solving the Helmholtz equation with BumpNet-PINN. \label{fig-helmotz-pred}}
\end{figure}

Bump-PINN achieves an accurate prediction with only $280$ parameters (\cref{fig-helmotz-pred}). In contrast, PINN requires approximately $55$ times more parameters to achieve high accuracy, that requires fewer parameters than deep neural network (\cref{tbl:benchmark-pde}). SPINN achieves similar accuracy and efficiency with 400 nodes.

\subsubsection{Two-Dimensional Poisson Equation}
\label{sec:poisson2d}

\begin{figure}[h]
  \begin{subfigure}{\linewidth}
      \includegraphics[width=\linewidth]{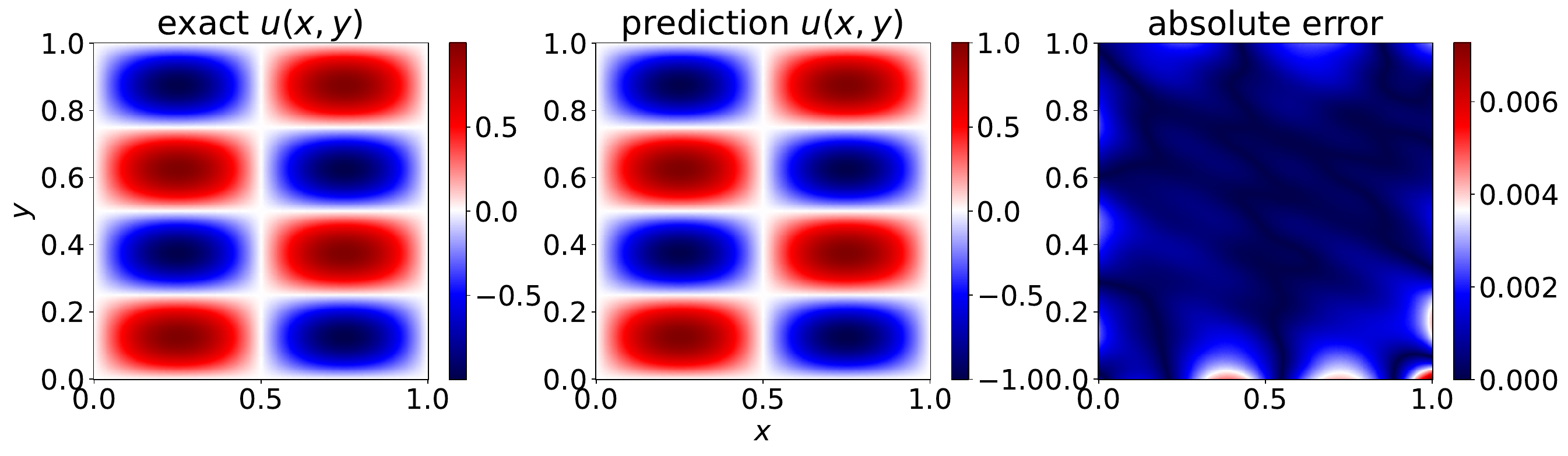}
      \subcaption{(\textit{Left}) Exact solution. (\textit{Middle}) Prediction of the BumpNet PINN. (\textit{Right}) Absolute difference.}
  \end{subfigure}
  \begin{subfigure}{\linewidth}
      \includegraphics[width=\linewidth]{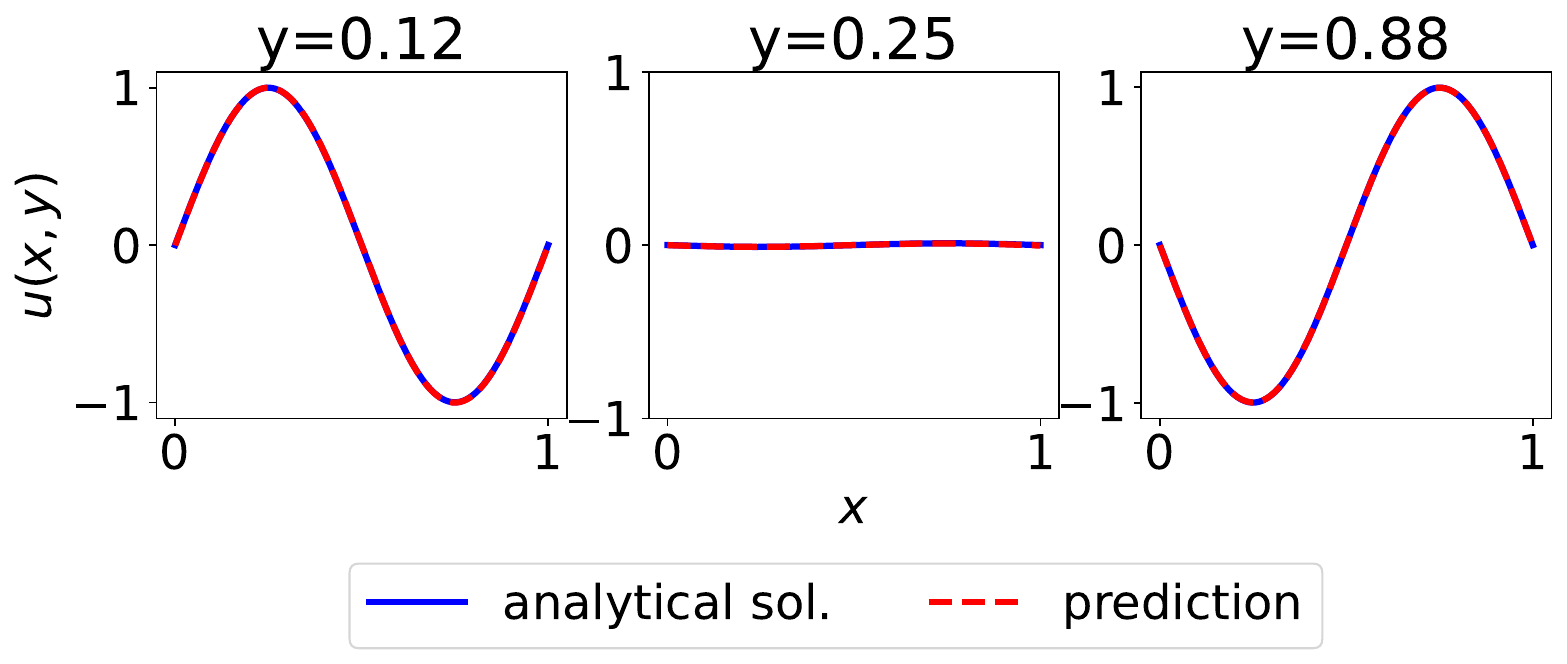}
      \subcaption{Comparison of prediction and exact solution on $x, y\in [0,1]^2$.}
  \end{subfigure}
  \caption{Solving the Poisson equation with BumpNet-PINN.}
  \label{fig-poisson}
\end{figure}

We now examine the application of Bump-PINN to the two-dimensional Poisson equation

\begin{equation}
%\begin{aligned}
    \frac{\partial^2 u(x,y)}{\partial x^2} + \frac{\partial^2 u(x,y)}{\partial y^2} \,=\, 20 \pi^2 \sin(2\pi x) \sin(4\pi y)\,,
%    x, y &\in \Omega = (0,1) \times (0,1), \\
%    u(x,y) &= 0, \quad x, y \in \partial \Omega,
%\end{aligned}
\end{equation}
where $(x,y) \in D=[0,1]^2$, $u(x,y)=0$ for $ x,y\in \partial D$ for  zero boundary conditions. The solution is 
\begin{equation}
    u(x,y) = \sin(2\pi x)\sin(4\pi y)
\end{equation}
For this problem, Bump-PINN is initialized with a structured grid of $6 \times 6$ bumps on the two-dimensional plane without gaps. The initial learning rate is 0.5. Bump-PINN achieves a mean squared error (MSE) of $9.61 \times 10^{-4}$ on the test set, which is slightly less accurate than SPINN. However, Bump-PINN demonstrates remarkable computational efficiency, completing training in around 1 minute, which is {\color{black} 20x faster than SPINN} (\cref{fig-poisson} and \cref{tbl:benchmark-pde}). Both Bump-PINN and SPINN achieve similar accuracy with fewer parameters than the baseline PINN counterpart.

\subsubsection{Heat Equation\label{sec:heat1d}}

This benchmark evaluates Bump-PINN in a simple time-evolution problem, where BumpNet is trained on both spatial and temporal domains simultaneously, i.e., with ``space-time'' bumps. The benchmark is the 1D heat equation

\begin{equation}
\frac{\partial u(x,t)}{\partial t} \,=\, \alpha \frac{\partial^2 u(x,t)}{\partial x^2} + 2\sin(\pi x)\,, 
\end{equation}
where $(x,t) \in [0,1]\times [0,5]$, with initial condition $u(x,0) = \sin(2\pi x)$.
The analytical solution to this equation is given by
\begin{equation}
u(x,t) = e^{-4\pi^2 \alpha t}\sin(2\pi x) + \frac{2}{\pi^2 \alpha}(1 - e^{-\pi^2 \alpha t})\sin(\pi x)\,,
\end{equation}
where we set $\alpha=1$. 

\begin{figure}[h]
  \begin{subfigure}{\linewidth}
      \includegraphics[width=\linewidth]{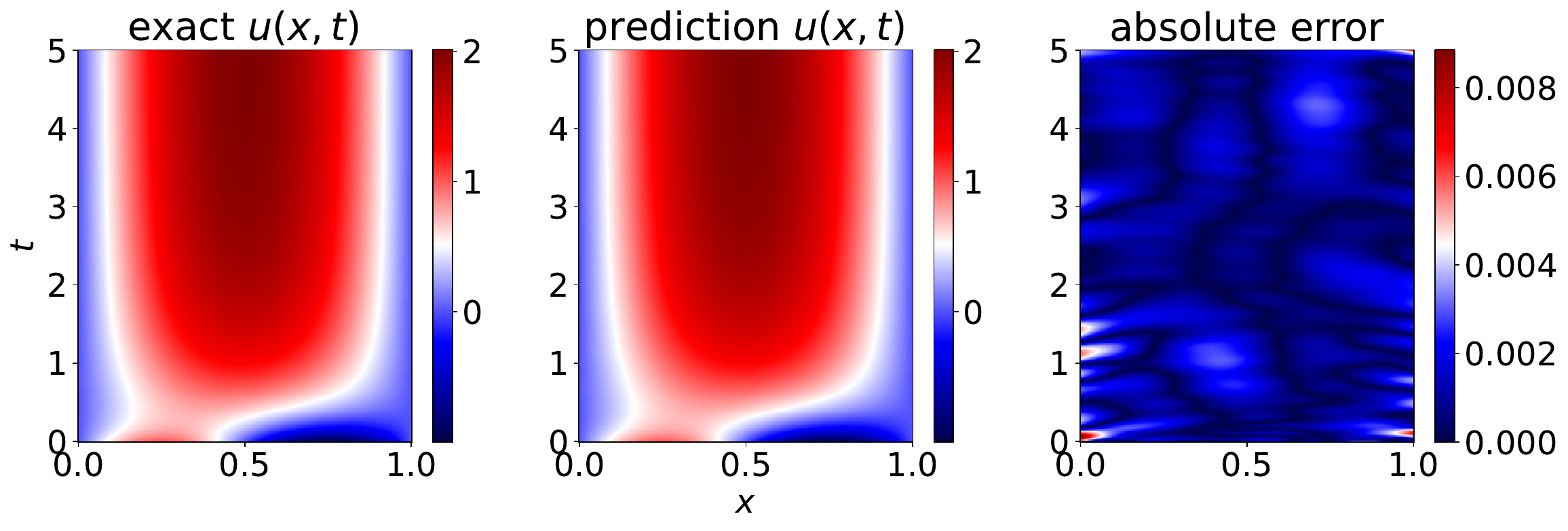}
      \subcaption{(\textit{Left}) Exact solution. (\textit{Middle}) Prediction of the Bump-PINN. (\textit{Right}) Absolute difference.}
  \end{subfigure}
  \begin{subfigure}{\linewidth}
      \includegraphics[width=\linewidth]{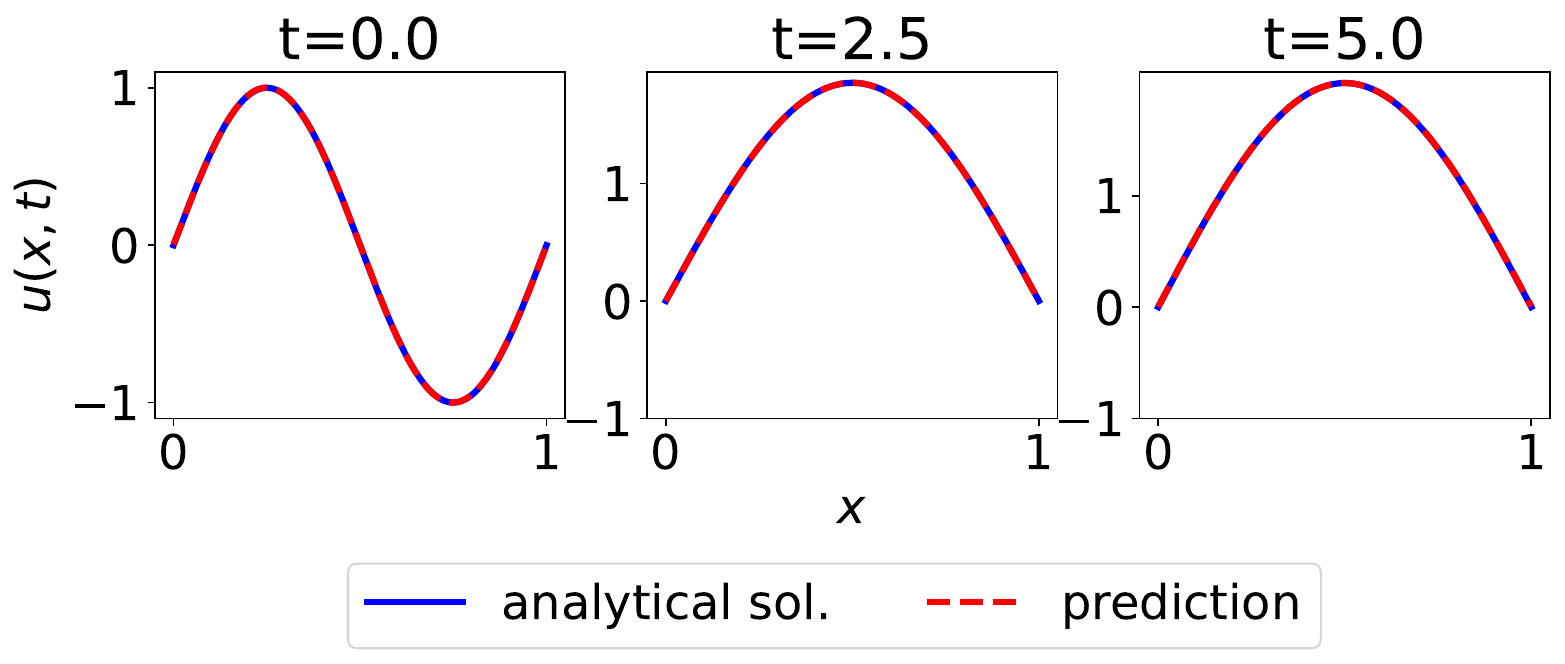}
      \subcaption{Comparison of preciction and exact solution on $t= [0, 5.0]$ and $x\in [0,1]$.}
  \end{subfigure}
  \caption{Solving the Heat equation with Bump-PINN.}
  \label{fig-heat1d}
\end{figure}

We applied Bump-PINN with $20 \times 6$ structured bumps along the spatial and temporal dimensions, resulting in 840 parameters. We use higher resolution on the x-axis to capture the sinusoidal curve. The Bump-PINN was compared to a PINN based on a MLP with 15 layers of 15 nodes each, containing 3,421 parameters, and SPINN with 200 nodes and 1,021 parameters. Despite its smaller parameter count, Bump-PINN demonstrated superior predictive accuracy in capturing complex solution geometries (Fig.~\ref{fig-heat1d} and Table~\ref{tbl:benchmark-pde}). This efficiency underscores its architectural advantages. Notably, SPINN faces convergence challenges for this problem and requires supplemental finite difference methods to achieve an accurate solution~\cite{ramabathiran2021}. In contrast, BumpNet is flexible and can effectively represent intricate solution geometries without additional numerical methods.

\subsubsection{Advection Equation with Periodic Boundary
Condition}\label{sec:advection}

Another example of a space-time problem is the 1D advection equation

% \begin{equation}
%   \begin{aligned}
%     \frac{\partial u(x,t)}{\partial t} + \nu\frac{\partial u(x,t)}{\partial x} &= 0, \quad (x,t) \in [0,2\pi] \times [0,T]\\
%     u(0,t) &= u(2\pi, t), \quad t \in [0,T]\\
%     u(x,0) &= \sin(x)
%   \end{aligned}
% \end{equation}

\begin{equation}
\frac{\partial u(x,t)}{\partial t} + \nu\frac{\partial u(x,t)}{\partial x} \,=\, 0\,,
 \end{equation}
where $x\in [0,2\pi], t \in [0,1]$ with initial condition $u(x,0) = \sin(x)$ and periodic boundary condition $u(0,t) = u(2\pi,t)$. The solution is 
\begin{equation}
    u(x,t) = \sin(x-\nu t)
\end{equation}

In this experiment, we set $\nu=30$, a regime where PINN fails to train~\cite{mcclenny2023self, braga-neto_characteristics-informed_2022}. The bumps in BumpNet are initialized on a $(2 \times 11)$ grid for x-axis and t-axis. A self-adaptive loss function with a square mask is employed, where the self-adaptive weights are initialized to 1.0. Both BumpNet and the self-adaptive weights are optimized using the Adam optimizer with an initial learning rate of 0.02. Under these settings, the L1 error achieved is $2.3 \times 10^{-3}$.  Furthermore, similar to PINN, Bump-PINN also fails to train when $\nu=30$. This experiment highlights the efficiency of BumpNet compared to MLP within the self-adaptive framework.

\begin{figure}[h]
  \begin{subfigure}{\linewidth}
      \includegraphics[width=\linewidth]{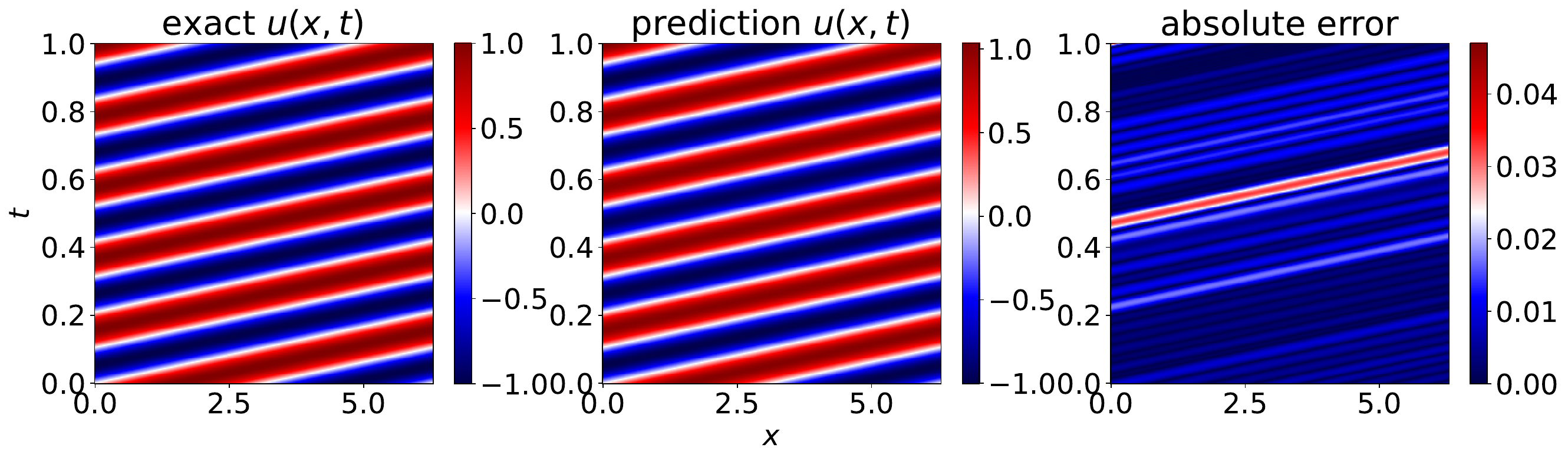}
      \subcaption{(\textit{Left}) Exact solution. (\textit{Middle}) Prediction of the Bump-PINN. (\textit{Right}) Absolute difference.}
  \end{subfigure}
  \begin{subfigure}{\linewidth}
      \includegraphics[width=\linewidth]{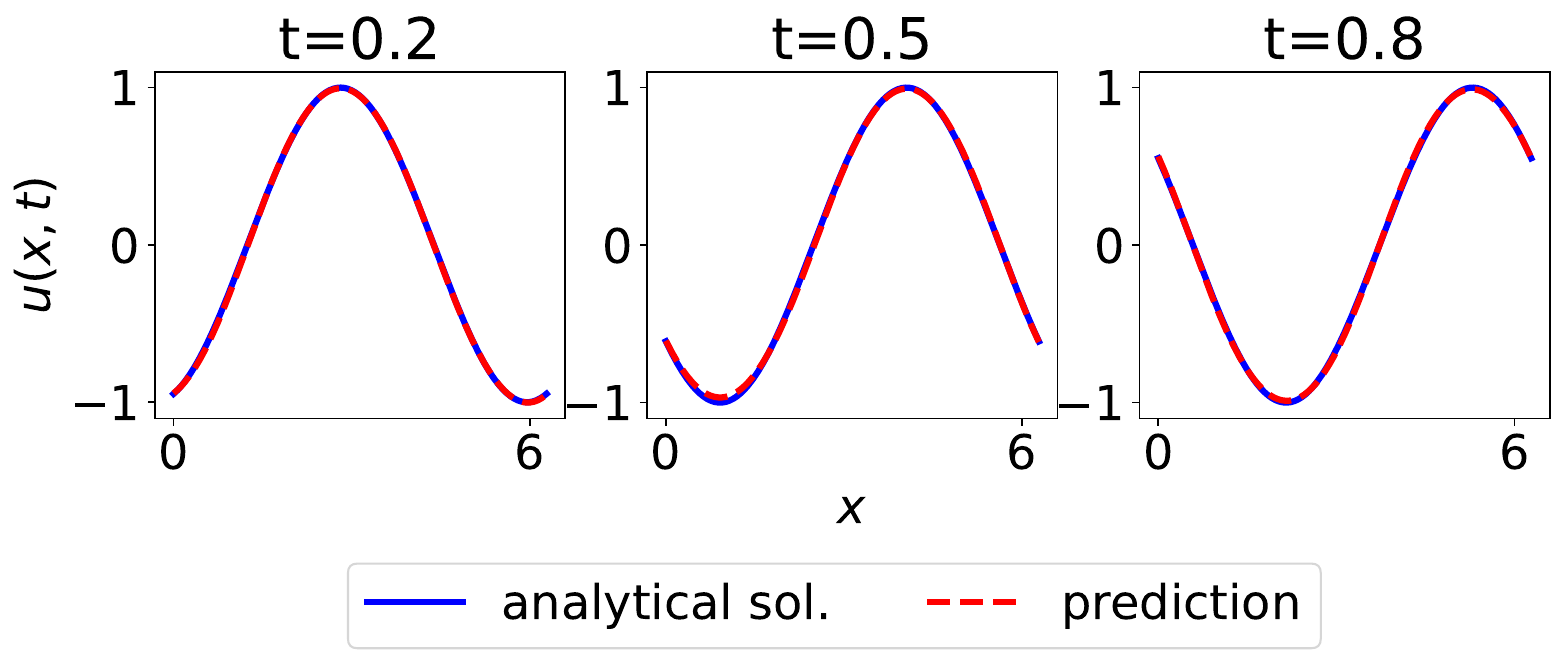}
      \subcaption{Comparison of preciction and exact solution on $t= -1.5, 0.0, 1.5$ over $x\in [0,6]$.}
  \end{subfigure}
  \caption{Solving Linear advection equation with Bump-SAPINN.}
\end{figure}

\vspace{2ex}

\begin{table*}
    \centering
    \begin{tabular}{llrrr}
        \toprule
        benchmark & method & L1 error  & \# param & train time (sec) \\
        \midrule
        Advection Equation & Bump-PINN & 2.30e-03 & {\bf 154} & {\bf 28} \\
         (60K epochs) & Bump-SAPINN & 8.04e-03 & {\bf 154} & 113 \\
         & PINN & 7.78e-02 & 83073 & 48  \\
         & SAPINN & {\bf 3.07e-04} & 83073 & 727  \\
         & SPINN & 6.33e-01 & 481 & 437  \\
        \midrule
        Heat Equation & Bump-PINN & {\bf 1.83e-03} & {\bf 840} & {\bf 12}  \\
         (20K epochs) & Bump-SAPINN & 6.51e-03 & {\bf 840} & 111 \\
         & PINN & 3.15e-02 & 3421 & 41  \\
         & SAPINN & 4.56e-02 & 3421 & 144  \\
         & SPINN & 2.18e+00 & 1021 & 291 \\
        \midrule
        Helmholtz Equation & Bump-PINN & 2.66e-02 & {\bf 1575} & {\bf 51}  \\
        % & Bump-PINNPrune & 6.91e-03 & 1470 & 73  \\
         (70K epochs) & Bump-SAPINN & {\bf 7.92e-03} & {\bf 1575} & 406  \\
          & PINN & 1.24e-01 & 83073 & 85  \\
          %& PINNBumpPrune0 & 1.64e-02 & 2800 & 59  \\
          & SAPINN & 5.04e-02 & 83073 & 1283  \\
          & SPINN & 9.83e-01 & 1761 & 2265  \\
        \midrule
        Poisson Equation & Bump-PINN & 8.66e-04 & {\bf 252} & {\bf 61}  \\
        (100K epochs) & Bump-SAPINN & 7.45e-03 & {\bf 252} & 79  \\
         & PINN & 2.73e-02 & 83073 & 112  \\
         & SAPINN & 2.78e-02 & 83073 & 1600  \\
         & SPINN & {\bf 1.03e-04} & 481 & 1256  \\
        \bottomrule
    \end{tabular}
    \caption{Comparison of PINN, SAPINN, Bump-PINN, Bump-SAPINN, and SPINN.}
    \label{tbl:benchmark-pde}
\end{table*}

\begin{table*}
    \centering
    \begin{tabular}{llrr}
       \toprule
        benchmark & method & \# collocation points & \# boundary points\\ 
       \midrule
        Helmholtz & PINN  & 1000 & 200\\
         & SAPINN  & 10000 & 200\\
         \midrule
        Poisson & PINN & 1000 & 100\\ 
        & SAPINN  & 10000 & 200\\ 
        \midrule
        Heat & PINN & 1000 & 200\\
        & SAPINN  & 20000 & 1000\\ 
        \midrule
        Advection & PINN & 1000 & 200\\ 
         & SAPINN  & 15000& 1000\\
        \bottomrule
    \end{tabular}
    \caption{Experimental Settings. ``PINN'' refers to the baseline PINN, Bump-PINN, and SPINN, while ``SAPINN'' refers to the baseline SAPINN and the Bump-SAPINN. The number of boundary points is per side.}
\end{table*}

\subsection{Performance of Bump-PINN with Pruning}\label{sec:res-pruning}

\begin{figure}
    \includegraphics[width=0.9\linewidth]{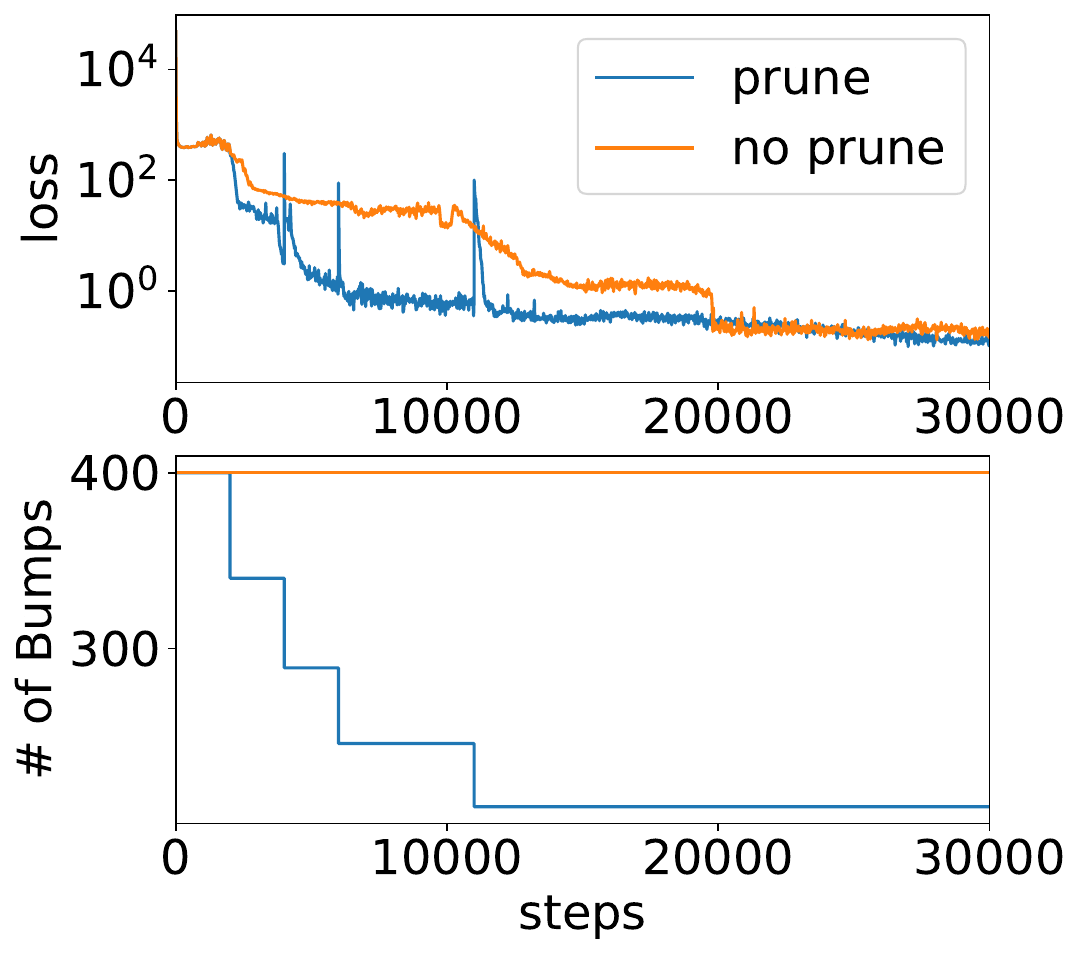}
    \caption{Training a Bump-Net with and without pruning.}
    \label{fig:prune-cmp}
\end{figure}

To investigate the possibility of performing automatic model order reduction with BumpNet (see Section~\ref{sec:method-pruning}), we compare the performance of Bump-PINN {\em with} and {\em without} pruning in the Helmholtz problem (\cref{eq-helmotz-prob}). Bumps with $0.15\%$ lowest absolute amplitude are pruned for 4 times with each interval contained 2000 iterations. Since the pruned model has a different number of parameters, the optimizer is re-initialized for the pruned model, which introduces a spike in the training loss right after pruning, as seen in \cref{fig:prune-cmp}. Nevertheless, we can see in this figure that pruning accelerates training convergence over the baseline Bump-PINN without pruning. This suggests that removing non-essential bumps modifies the loss surface in a favorable way.

\subsection{Bump-EDNN Performance}\label{sec:res-evo}

\begin{figure}
    \begin{subfigure}[t]{0.25\textwidth}
        \includegraphics[width=\textwidth]{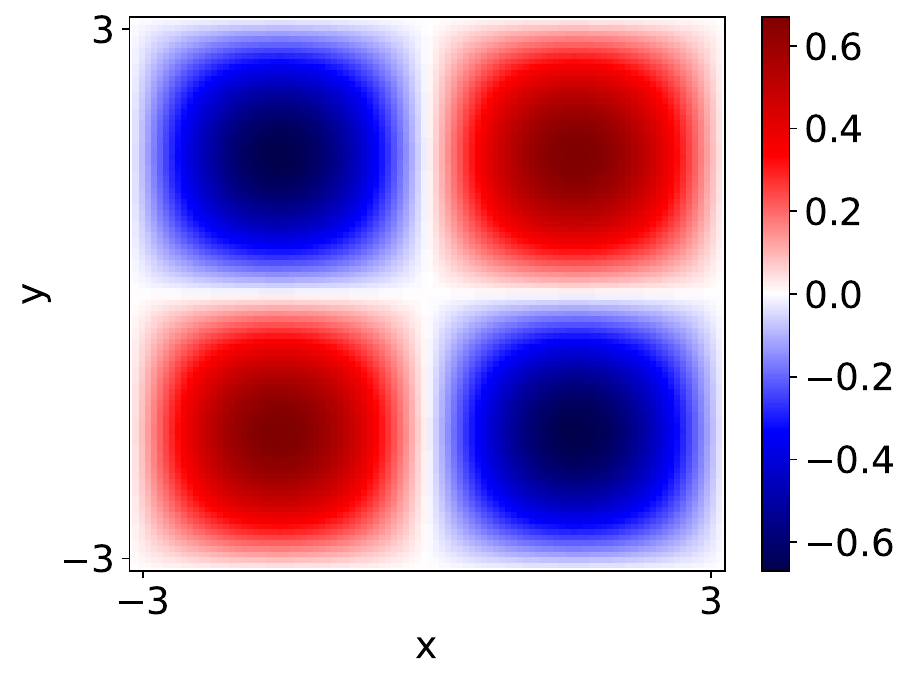}
        \subcaption{Analytical solution}
    \end{subfigure}%
    \begin{subfigure}[t]{0.25\textwidth}
        \includegraphics[width=\textwidth]{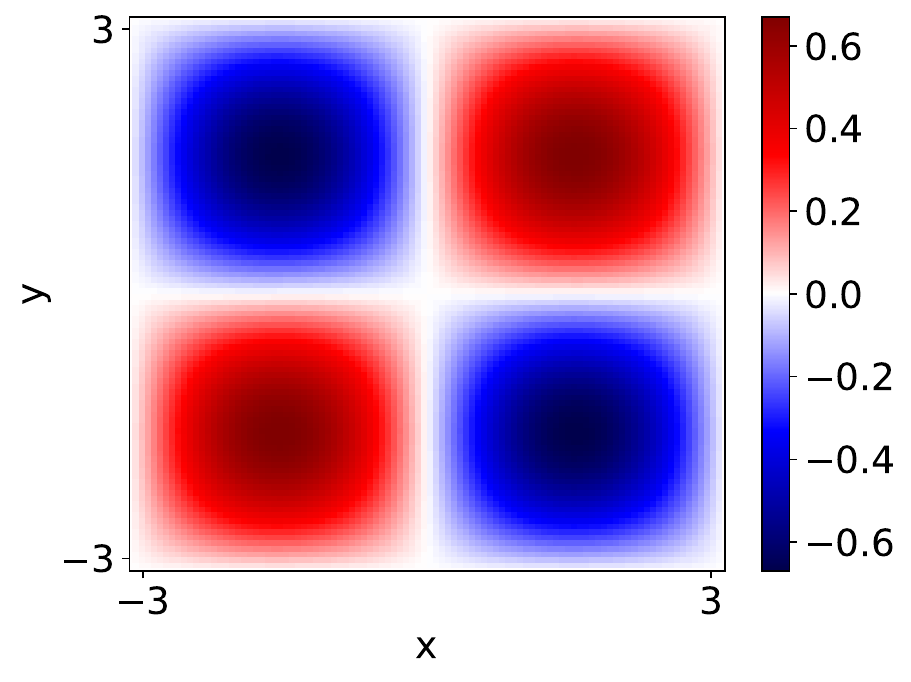}
        \subcaption{Evolutional Bump prediction}
    \end{subfigure}
     \begin{subfigure}[t]{0.25\textwidth}
        \includegraphics[width=\textwidth]{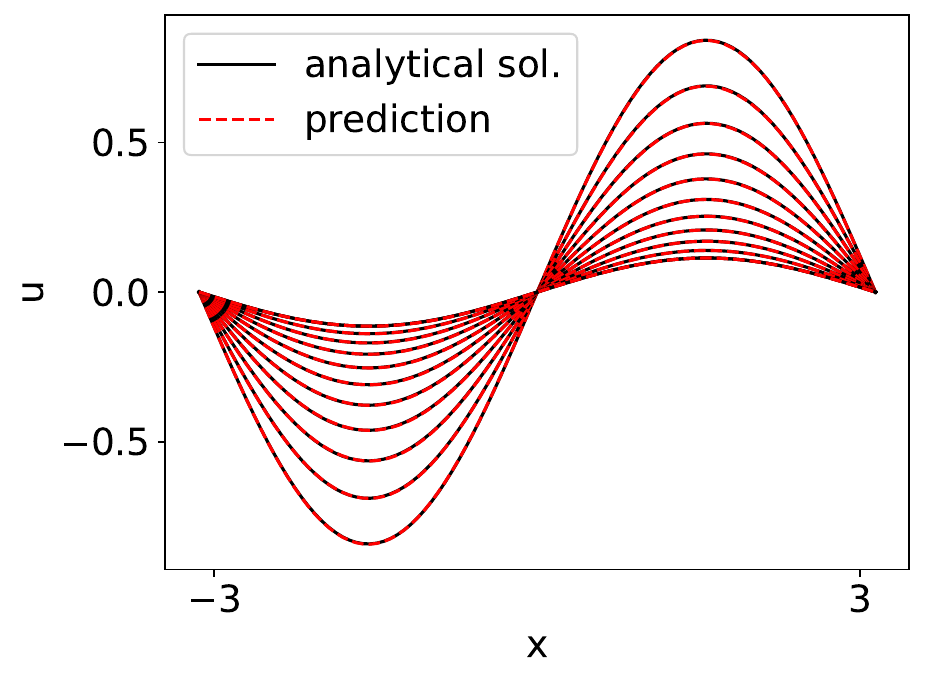}
        \subcaption{1D profile comparison}
    \end{subfigure}%
     \begin{subfigure}[t]{0.25\textwidth}
        \includegraphics[width=\textwidth]{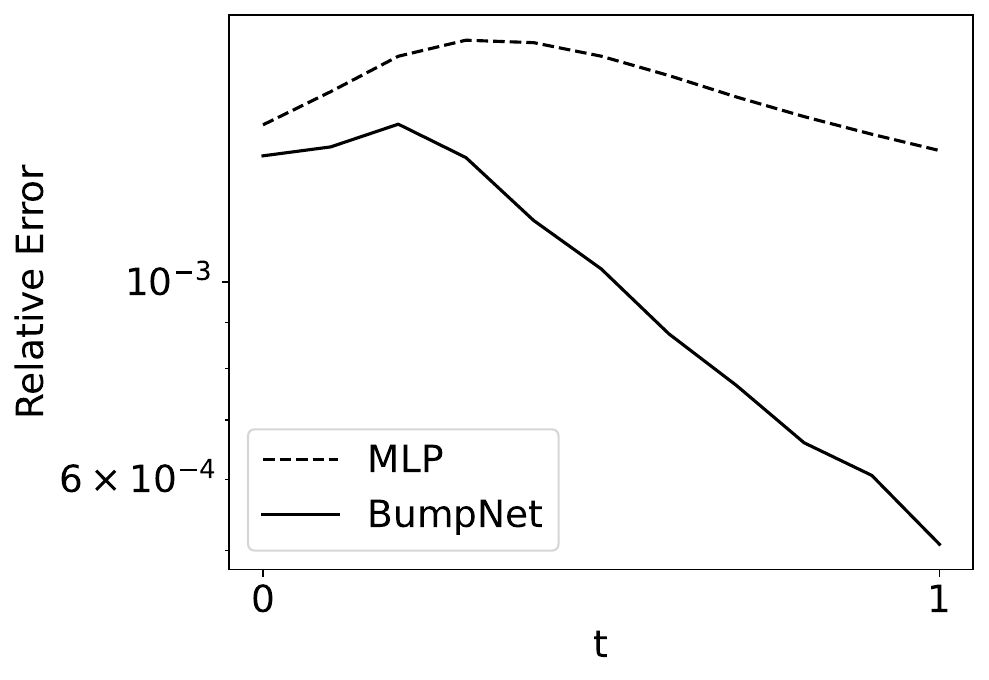}
        \subcaption{Relative Error versus time}
    \end{subfigure}
    \caption{Solving the Heat equation with Bump-EDNN. (a,b) Comparison between prediction and analytical solution at $t=0.2$. (c) 1D profile comparison at $y=1$. (b) Relative error of Bump-EDNN compared to EDNN.}
    \label{fig:evo-heat}
\end{figure}

In this section, we demonstrate the advantages of Bump-EDNN in comparison to the standard EDNN in solving time-dependent PDEs with significantly reduced computational cost and improved efficiency. We apply this approach on the two-dimensional heat equation defined on $\Omega = [-\pi, \pi]^2$
\begin{equation}
    \begin{split}
        \frac{\partial u(x,y,t)}{\partial t} &= \left(\frac{\partial^2 u(x,y,t)}{\partial x^2} + \frac{\partial^2 u(x, y, t)}{\partial y^2}\right),\\
        u(x, y, 0) &= \sin(x) \sin(y)\,,\\
        u(x,y,t)&=0 \text{ on } \partial\Omega\,.\\
    \end{split}
\end{equation}
The analytical solution is 
\begin{equation}
    u(x,y,t) = \sin(x)\sin(y)\exp(-2 t)\,.
\end{equation}
We impose the boundary conditions on the neural networks using the approach in \cite{lu2021physics}.
%, and in the form of
%\begin{equation}
%5\begin{split}
%   \hat{u}(x,y;\theta) &= (1-e^{-\pi - x})(1-e^{x-\pi})\\
%                        &(1-e^{-\pi-y})(1-e^{y-\pi})NN(x,y;\theta)
%\end{split}
%\end{equation}
%where $NN$ is neural network to optimize with parameter $\theta$. 
In the experiment, the parameter $\gamma$ is solved for with tolerance 1e-4. The EDNN is based on an MLP with four layers and 20 neurons per layer. The ODE system for evolving the weights is solved by the third-order Runge-Kutta solver, for a good balance between accuracy and computational time.
We can see in \cref{fig:evo-heat} that Bump-EDNN is more accurate than EDNN, especially as time increases. We can see in Table~\ref{tbl:evo-bump-vs-mlp} that Bump-EDNN converged faster than EDNN during training on the initial condition, and that the higher accuracy of Bump-EDNN is obtained with a much smaller number of parameters and faster solving time than EDNN. In particular, the time required for weight evolution, once the weights have been initialized by training on the initial condition, is almost three orders of magnitude shorter in the case of Bump-EDNN.

\begin{table}
    \centering
    \begin{tabular}{ccc}
       & \textbf{Bump-EDNN} & \textbf{EDNN}\\
        \hline
    \textbf{Architecture} &6 $\times$ 6 bumps & $4 \times 20$ MLP\\ 
    \textbf{\# of trainable parameters} & 252 & 1341\\
    \textbf{\# of evolving parameters} & 36 & 1341\\
    \textbf{Training time} & 6m43s & %(99.15 iter/s) &
    8m24sec \\ %(79.24 iter / s)\\
    %\textbf{L2 Error at initial} & 2.79e-4 & 2.44e-6\\
    \textbf{L2 Error at $t=1$}  & 5e-4 & 1.4e-3\\
    \textbf{Time-evolution solving time}  & 6sec & 23m13sec\\\hline
    \end{tabular}
    \vspace{1ex}
    \caption{Comparison of Bump-EDNN and EDNN.}
    \label{tbl:evo-bump-vs-mlp}
\end{table}

\subsection{Bump-DeepONet Performance}\label{sec:res-deeponet}

To investigate Bump-DeepONet's potential for operator learning, we consider a nonlinear diffusion-reaction PDE with a forcing term $f(x)$:
\begin{equation}
    \frac{\partial u}{\partial t} = D\frac{\partial^2 u}{\partial x^2} + ku^2 + f(x), \quad (x,t) \in (0,1] \times (0,1],
    \label{eq:rd}
\end{equation}
where the reaction rate is $k=0.01$ and the diffusion coefficient is $D=0.01$. Zero initial and boundary conditions are used. The operator learning goal is to recover the map from the forcing term $f(x)$ to the corresponding solution $u(x,t)$. Following~\cite{wang2021learning}, the forcing function $f(x)$ is generated using a Gaussian Random Field (GRF) with an square exponential kernel, a length scale of 0.2, and a scale factor of 1.

\begin{figure}[h]
  \begin{subfigure}{\linewidth}
      \includegraphics[width=\textwidth]{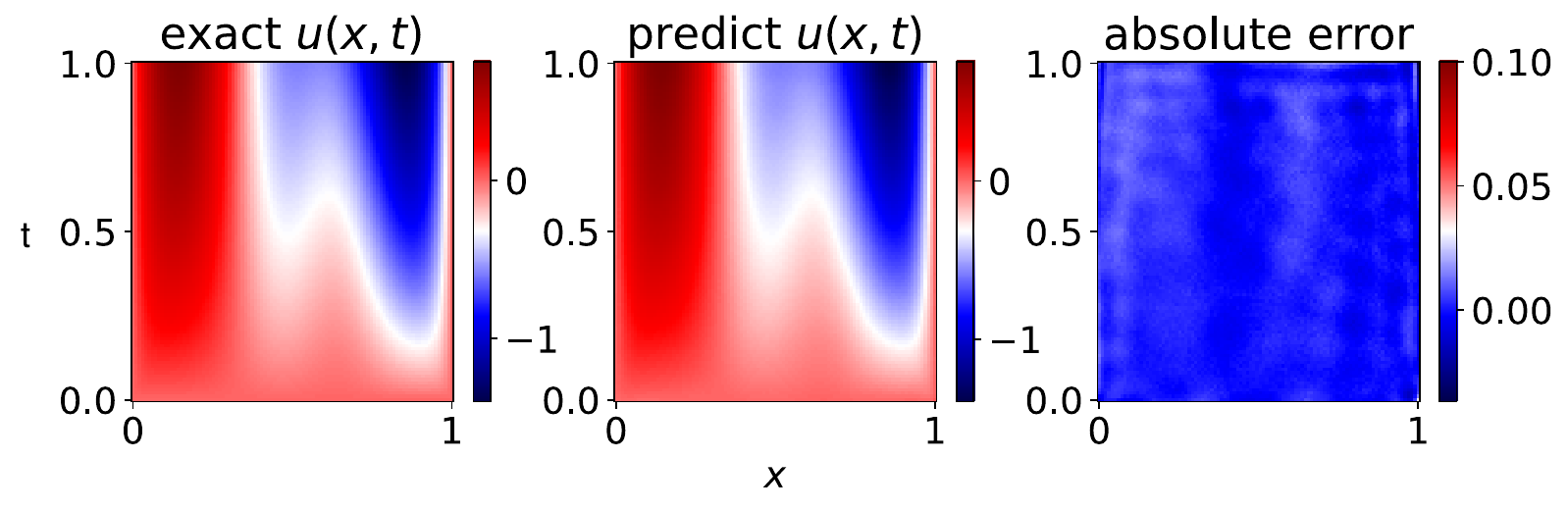}
      \subcaption{Solution for a random test forcing function.}
  \end{subfigure}
  \begin{subfigure}{\linewidth}
      \includegraphics[width=0.95\textwidth]{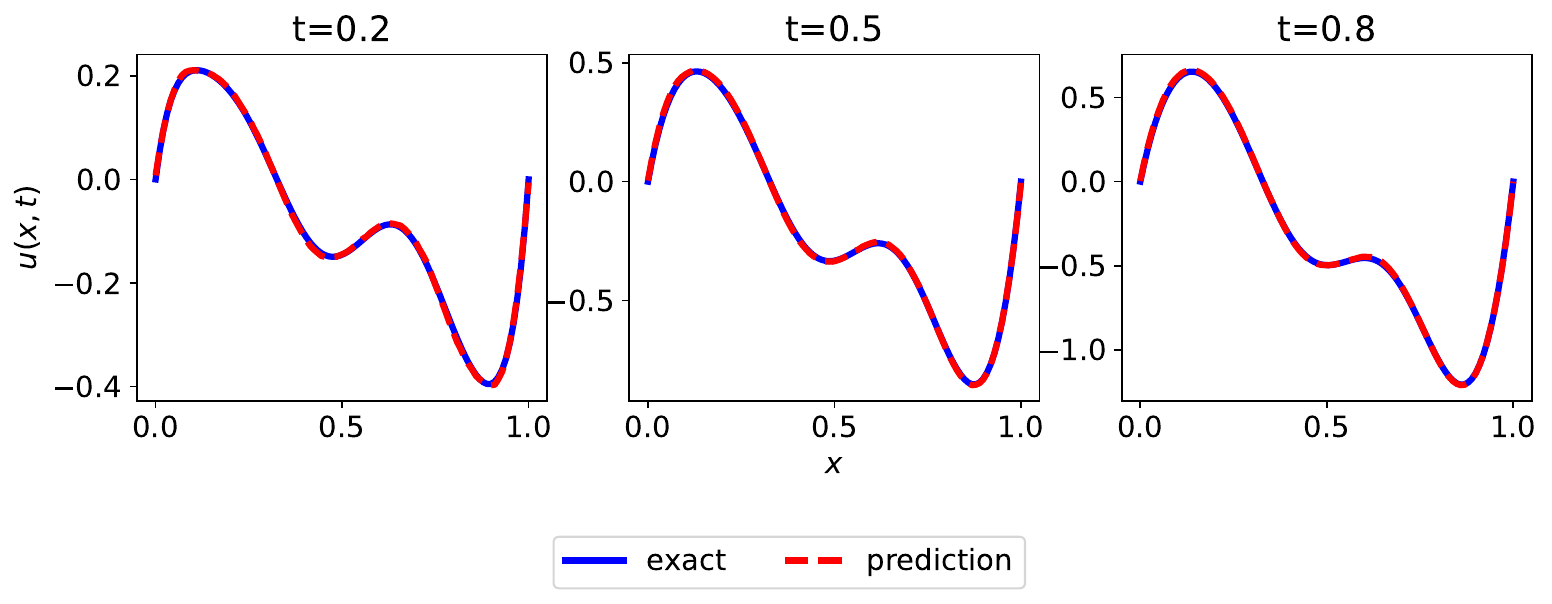}
      \subcaption{Solution in (a) at different time slices.}
  \end{subfigure}
  \begin{subfigure}{\linewidth}
    \includegraphics[width=0.96\textwidth]{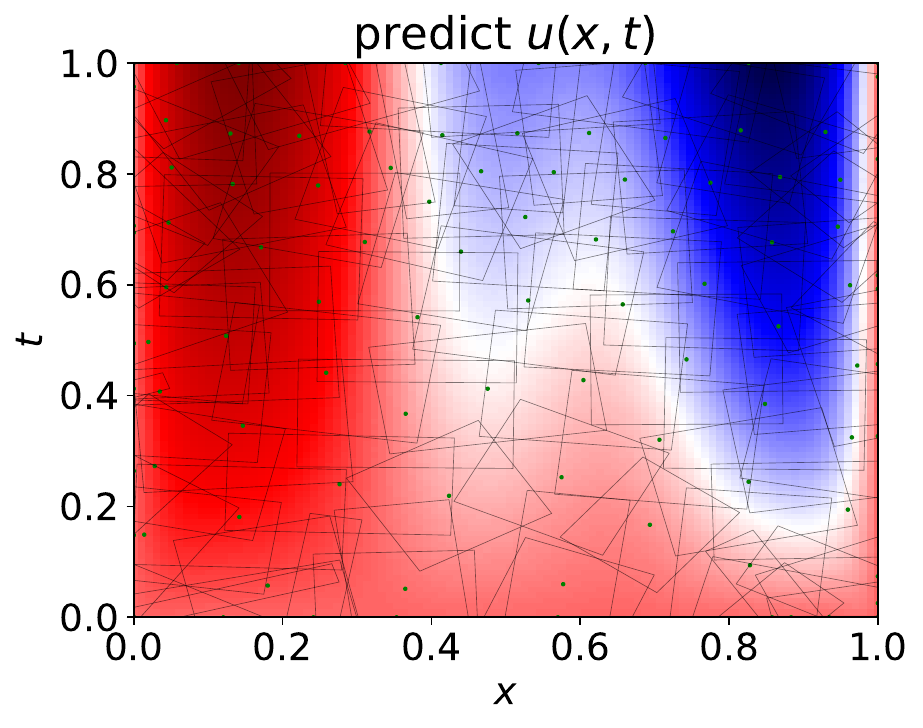}
    \subcaption{Bump locations and boundaries for the solution in (a).\label{fig-deeponet-diffusion-vis}}
  \end{subfigure}
  \caption{Solving a diffusion reaction PDE with Bump-DeepONet. In part (c), the dots indicate bump centers, while the rectangles represent bump bouundaries.\label{fig-deeponet-diffusion}}
\end{figure}

For this task, we apply a $10 \times 10$ grid of bumps across the spatial and temporal domains. Bumps are constrained to be in the physical domain, as explained in Section~\ref{sec:methods}.\ref{sec:detailed}, for efficient representation. We observed that Bump-DeepONet accurately solves the operator learning problem, as can be seen in \cref{fig-deeponet-diffusion} for a random test forcing function (not seen during training). Notably, Bump-DeepONet requires 100 times fewer parameters than the DeepONet while achieving comparable accuracy in this problem, as seen in \cref{tbl:deeponet-bumpnet-vs-mlp}. The trained Bump-DeepONet identifies the solution accurately by placing more bumps over regions of high solution variability (\cref{fig-deeponet-diffusion-vis}). These results demonstrate that BumpNet enables efficient operator learning and serves as a highly parameter-efficient alternative to traditional MLP-based architectures.

\begin{table}
    \tiny
    \centering
    \begin{tabular}{ccc}
        & \textbf{Bump-DeepONet} & \textbf{DeepONet}\\
        \hline
    \textbf{Architecture of trunk} & 10 $\times$ 10 & 50 wide, 4 layers\\
    \textbf{Training time} & 86 sec (937.2 iter/sec) & 158 sec (775.75 iter/sec) \\ 
    \textbf{\# of param. in trunk} &  600 & 25600\\ 
    \textbf{Test Error} & 8.12e-6 & 5.42e-6\\ \hline
    \end{tabular}
    \vspace{1ex}
    \caption{Comparison of Bump-DeepONet with DeepONet.}
    \label{tbl:deeponet-bumpnet-vs-mlp}
\end{table}

\section{Conclusion}

We introduced BumpNet, a parameter-efficient, fully-interpretable MLP architecture for learning PDE solutions. Extensive experiments demonstrated that Bump-PINN, Bump-EDNN, and Bump-DeepONet obtain accurate solutions in physics-informed machine learning and operator learning problems, while requiring fewer parameters, shorter training time, and faster predictions than competing methods. A simple pruning technique based on bump amplitudes accelerates training convergence. For time-dependent partial differential equations (PDEs), BumpNets can be applied via a Bump-PINN over the entire spatio-temporal domain, or by applying a Bump-EDNN, which only requires training a spatial BumpNet on the initial condition, and evolving its parameters using the evolutional neural network framework, which is fast, while maintaining low error during time-stepping. Bump-DeepONet allows the application of BumpNets to operator learning by replacing the trunk network of a DeepONet with a BumpNet. This efficient parameterization reduces the size of the trunk by half and accelerates convergence when learning an operator. We proved that BumpNets and Bump-DeepONets
are universal approximators of continuous functions and continuous operators, respectively. We believe that the efficiency and interpretability of the BumpNet architecture make it a promising new tool for advancing scientific machine learning and complex system modeling.

\section{Supplementary Material}
%Supplementary material is available at PNAS Nexus online.

We provide below the proofs of Theorems 1 and 2. The following lemma will be required. 

\vspace{1ex}
\noindent
{\bf Lemma.} %[Rectangle approximation on Lipschitz domains]
Let $\Omega \subset \mathbb{R}^d$ be a compact domain, and let
$f \in C(\Omega)$. For any $\varepsilon > 0$, there exists a finite collection of hyperrectangles
$\{R_k\}_{k=1}^N \subset \mathbb{R}^d$
and constants $c_k \in \mathbb{R}$ such that the function
\[
g(\v{x}) \,=\, \sum_{k=1}^N c_k \mathbf{1}_{R_k \cap \Omega}(\v{x})
\]
satisfies
\[
|\, f(\v{x}) - g(\v{x})\,| \,<\, \varepsilon\,, \:\: \textrm{for all} \:\: \v{x} \in \Omega.
\]

\begin{proof}
Since $\Omega$ is compact and $f$ is continuous, $f$ is uniformly continuous on $\Omega$, by the Heine-Cantor Theorem.  
Thus, for the given $\varepsilon > 0$, there exists $\delta > 0$ such that, for $\v{x},\v{y} \in \Omega$,
\[
|f(\v{x}) - f(\v{y})| \,<\, \varepsilon\,,\:\: \text{whenever} \:\:
\|\v{x}-\v{y}\| \,<\, \delta.
\]
Cover $\Omega$ by $N$ disjoint hyperrectangles $\{R_k\}_{k=1}^{N}$ such that
\[
\mathrm{diam}(R_k) \,:= \sup_{\v{x},\v{y}\in R_k} \|\v{x}-\v{y}\| \,<\, \delta\,.
\]
For each $R_k$, choose $\v{x}_k \in R_k \cap \Omega$ and set $c_k = f(\v{x}_k)$.
Let 
\[
g(\v{x}) \,=\, \sum_{k=1}^N c_k \mathbf{1}_{R_k \cap \Omega}(\v{x})\,.
\]
For all $\v{x} \in \Omega$, there exists $k$ such that $\v{x} \in R_k \cap \Omega$, so that
$\|\v{x} - \v{x}_k\| < \delta$, which in turn implies that
\[
|f(\v{x}) - g(\v{x})| \,=\, |f(\v{x}) - f(\v{x}_k)|\,<\, \varepsilon\,,
\]
proving the claim.
\end{proof}

% \begin{theorem}
% Let $\Omega \subset R ^d$ be a compact physical domain. The class of BumpNets defined by equations \eqref{eq-Bump1}--\eqref{eq-Bump3} is dense in $C(\Omega)$ with respect to the uniform norm. That is, for any $f \in C(\Omega)$ and any $\varepsilon > 0$, there exists a BumpNet $\psi$ such~that
% \[
% \| f(\v{x}) - \psi(\v{x})\| < \varepsilon\,, \:\: \textrm{for all} \:\: \v{x} \in \Omega.
% \]
% \end{theorem}

% \begin{proof}
% A
% \end{proof}

\vspace{1ex}
\noindent
{\bf Proof of Theorem 1.}
Using the  Lemma, we know that there exists a finite collection of hyperrectangles $\{R_k\}_{k=1}^N \subset \mathbb{R}^d$ and constants $c_k \in \mathbb{R}$ such that
\[
\left|
f(\v{x}) - \sum_{k=1}^N c_k \mathbf{1}_{R_k \cap \Omega}(\v{x})
\right| \,
<\, \frac{\varepsilon}{2}\,, \:\: \textrm{for all} \:\: \v{x} \in \Omega.
\]
Now, it is easy to see that 
\[
  \lim_{p \to \infty} \tanh (px) \,=\, \textrm{sgn}(x)\,=\, \begin{cases}+ 1\,,& x\geq 0\,,\\ -1\,,& x<0\,, \end{cases}
\]  
so that each Bumpnet basis function $\psi_i(\v{x})$ in \eqref{eq-Bump2} converges pointwise to the indicator function of a hyperrectangle:
\[
\lim_{p_i \to \infty} \psi_i(\v{x}) \,=\, R_i \,=\, \bigcap_{j=1}^d
\left\{\v{x} \in R^d \mid 
-\bar{s}^{\,i}_j \,\le\, \v{x}^T\v{\beta}^i_j \,\le\, \bar{s}^{\,i}_j \right\}.
\]
with location, orientation, and side lengths determined by its parameters. Hence, 
by choosing the sharpness parameter sufficiently large,  we can obtain a BumpNet basis function $\psi_k$ such that
\[
|\psi_k(\v{x}) - \mathbf{1}_{R_k}(\v{x})| \,<\,
\frac{\varepsilon}{2 \sum_{k=1}^N |c_k|}\,,\:\: \textrm{for all} \:\: \v{x} \in \Omega\,,
\]
for each hyperrectangle $R_k$. Define
\[
\psi(\v{x}) \,=\, \sum_{k=1}^N c_k \psi_k(\v{x})\,.
\]
Then
\[
\begin{aligned}
&|\,f(\v{x}) - \psi(\v{x})\,| \\[1ex] &\leq\,
\left|\,f(\v{x}) - \sum_{k=1}^N c_k \mathbf{1}_{R_k}(\v{x})]\,\right|
\:+\:
\left|\,\sum_{k=1}^N c_k (\mathbf{1}_{R_k}(\v{x}) - \psi_k(\v{x}))\,\right| \\
&\,<\,
\frac{\varepsilon}{2}
+
\sum_{k=1}^N |\,c_k\,| \,|\,\psi_k(\v{x}) - \mathbf{1}_{R_k}(\v{x})\,| \,=\,
\frac{\varepsilon}{2} + \frac{\varepsilon}{2} \,=\,\varepsilon\,,
\end{aligned}
\]
for all $\v{x} \in \Omega$. This completes the proof.

% \begin{theorem}
% Let $K \subset \mathbb{R}^{d_1}$ and $D \subset \mathbb{R}^{d_2}$ be compact sets, and let
% $\mathcal{U} \subset C(K)$ be compact under the uniform norm. If $\mathcal{G} : \mathcal{U} \to C(D)$ is any continuous operator, then for~any $\varepsilon > 0$, there exists a Bump--DeepONet operator $\mathcal{G}_{\mathrm{BDO}}$ such~that 
% \[
% \sup_{f \in \mathcal{U}}
% \left\|
% \mathcal{G}(f) - \mathcal{G}_{\mathrm{BDO}}(f)
% \right\|_{\infty}
% < \varepsilon .
% \]
% \end{theorem}

\vspace{2ex}
\noindent
{\bf Proof of Theorem 2.} 
The proof combines Theorem 1 with the original theorem on the universal approximation property of DeepONets, namely, Theorem~5 in \cite{chen1995universal}. This theorem ensures that if 
$K$ and $\Omega$ are compact sets in $R^p$ and $R^d$, respectively, $\mathcal{P}$ is a compact set in $C(K)$, $G:\mathcal{P} \rightarrow C(\Omega)$ is a continuous operator, and $\sigma$ is a continuous non-polynomial function, then for any $\varepsilon > 0$, 
there are positive integers
$p,n,m$, constants $c_i^k$, $\xi_j^k$, $\theta_i^k \in \mathbb{R}$, $w_k \in \mathbb{R}^d$,
$\v{x}_j \in K$, for $i=1,\ldots,n$, $k=1,\ldots,p$, and $j=1,\ldots,m$, such that
\begin{equation}
  %\begin{aligned}
  %  &
  \Bigg|
G(f)(\v{x}) 
 -\!
\sum_{i,k=1}^{n,p}
%\sum_{i=1}^n
\!c_i^k
\underbrace{
\sigma\!\left(
\sum_{j=1}^m
\xi_j^k\, f(\v{x}_j)
+
\theta_i^k
\!\right)
}_{\text{branch}}
\underbrace{
\sigma\!\left(
w_k \!\cdot\! \v{x} + \zeta_k
\right)
}_{\text{trunk}}
\Bigg|%\\
%& \quad <\, \varepsilon \,,
%\end{aligned}
\end{equation}
is at most $\varepsilon$, for all $f \in \mathcal{P}$ and $\v{x} \in \Omega$. By Theorem 1, each output of the the trunk network in the previous expression can be approximated to the required precision by a Bump-Net. This proves the theorem.

\section{Funding}
The work of STC and UBN was partially supported by the National Science Foundation under Grant Number 2225507, while IGK was partially supported by the US Department of Energy.

\section{Author contributions statement}
%Must include all authors, identified by initials, for example:
IGK and UBN provided conceptualization; STC conducted the experiments, analysed results. UBN and STC wrote the manuscript; UBN secured funding; all authors reviewed and approved the manuscript.

% \section{Previous presentation}
% %These results were previously presented at [conference, date].
% No previous presentation.

% \section{Preprints}

% There is a preprint at \url{https://arxiv.org/abs/2512.17198}.

% \section{Data availability}
% %The data underlying this article are available in [repository name, eg, the GenBank Nucleotide Database] at [URL], and can be accessed with [unique identifier, eg, accession number, deposition number].
% No database was used.

\bibliographystyle{icml2025}
\bibliography{ref}

\end{document}